\pdfoutput=1
\documentclass[10pt]{article}
\usepackage[a4paper,margin=22mm]{geometry}
\usepackage[T1]{fontenc}
\usepackage[utf8]{inputenc}
\usepackage{lmodern}
\usepackage{textcomp}
\usepackage{amsmath,amssymb}
\usepackage{graphicx,booktabs,tabularx,array,xcolor}
\usepackage[font=small,labelformat=empty,justification=raggedright,singlelinecheck=false]{caption}
\usepackage{microtype}
\usepackage{enumitem}
\usepackage[numbers,super,sort&compress]{natbib}
\usepackage[hidelinks]{hyperref}
\usepackage{url}
\setlist{nosep,leftmargin=*}
\definecolor{ink}{HTML}{1F2A44}
\definecolor{inksoft}{HTML}{4C5B78}
\definecolor{rose}{HTML}{D6336C}
\definecolor{tint}{HTML}{F2F5FB}
\newenvironment{thmbox}{\par\medskip\begingroup\leftskip=1em\rightskip=1em}{\par\endgroup\medskip}
\newenvironment{proofbox}{\par\smallskip\begingroup\small\leftskip=1em\rightskip=1em}{\par\endgroup\medskip}



\title{Sketchpad: certified exact arithmetic with a small vision-language model}
\author{Mostafa `Neo' Mohsenvand\\ {\small MIT Media Lab}}
\date{31 October 2026}
\begin{document}
\maketitle
\begin{abstract}
Language models are accurate on most long-arithmetic problems, but their accuracy is estimated on samples and carries no guarantee for an unseen input; tool use is exact but moves the computation outside the network. We show that exactness at all input lengths becomes a finite, checkable property of a network when each of its steps reads only a bounded view of an external working memory. In Sketchpad, Qwen3.5-0.8B with a low-rank adapter executes arithmetic procedures on a rendered sheet. Each forward pass reads a three-cell view and emits one step: a digit, a carry or borrow, and the next operation. A fixed runtime moves the view and writes no digit of a result. We prove in Lean that the model is then exact on all inputs of any length if and only if it is exact on the finite set of views the runtime presents, that no finite test suffices when steps also read earlier steps, and that the guarantee tolerates bounded numerical perturbation and extends to compositions of procedures. All 2,446 views pass on the deployment device, a laptop GPU running the model in a web browser. The certified model performs signed addition and subtraction, long multiplication and division, square roots and greatest common divisors at arbitrary length. It solved 200 of 200 pre-registered problems in the browser, and 220 of 220 sums of up to 1,000 digits and products of up to 20 digits, of which GPT-6.1 Sol, the strongest of ten other models, solved 217. A symbolic runtime built on these procedures solves, expands and factors polynomials with every number computed by the model, and a multiplicative gate on the adapter recovers the base model exactly when the sheet is not in use.

\end{abstract}
\section{Introduction}\label{sec-intro}
Language models now answer most long-arithmetic problems correctly, but this accuracy is measured on samples. Chain-of-thought prompting\cite{wei2022cot,nye2021scratchpad} and changes to positional encoding and data format\cite{nogueira2021limitations,anil2022length,lee2024teaching,zhou2024raspl,zhou2024robust,mcleish2024abacus,cho2024coupling} extend the lengths at which transformers add and multiply correctly. In each case, correctness on an unseen input remains an estimate, and the errors that remain are silent. In our comparison, GPT-6.1 Sol answered 217 of 220 problems correctly; its three errors, on sums of 300 and 1,000 digits, were completed answers well within the token budget, in the same form as the correct ones. Tool use\cite{schick2023toolformer,gao2023pal,chen2023pot,gou2024tora,feng2025retool} is exact, but places the computation outside the network.

A guarantee for all inputs requires that behaviour on unboundedly many inputs be fixed by finitely many facts. This holds when each step of a computation reads only a bounded view of an external memory and no other state passes between steps. The steps then range over a finite set of views, and a network that is correct on each view is correct on every run of any length. The observation is classical: it underlies Moore's analysis of experiments on finite automata\cite{moore1956gedanken}, the verification of learned recursive programs\cite{cai2017recursion} and a recent certificate for universal computation by language-model decoding\cite{lewandowski2026universal}. It also has a converse. If a step may read the steps before it, as in chain of thought, no finite test suffices.

\begin{figure}[tbp]
\centering
\includegraphics[width=\linewidth]{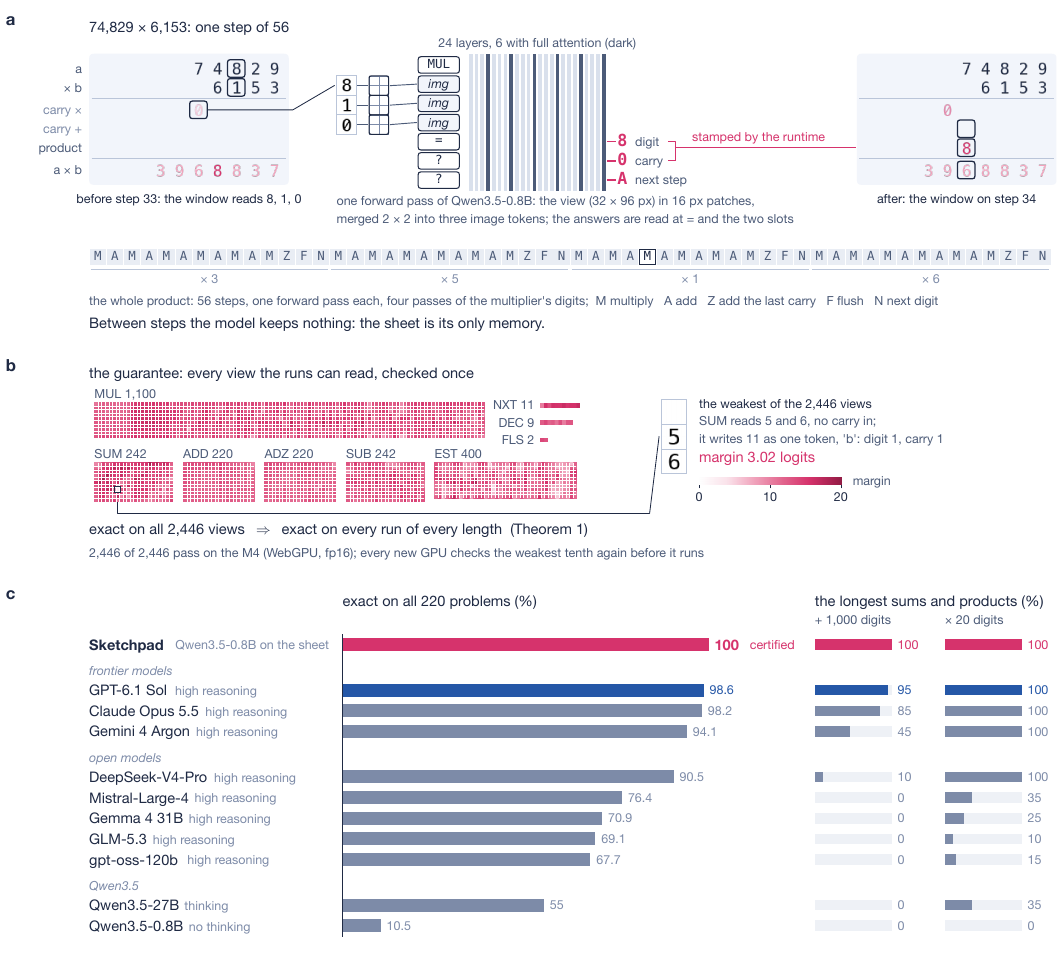}
\caption*{\textbf{Fig. 1 $|$} \textbf{Overview.} \textbf{a}, One step of the long multiplication 74,829 \ensuremath{\times} 6,153 (step 33 of 56). The view (three cells, 32 \ensuremath{\times} 96 pixels) enters Qwen3.5-0.8B as three image tokens after the operation token \texttt{MUL}; the outputs, read at \texttt{=\allowbreak{}} and two placeholder positions, are the digit, the carry and the next operation. Below: the operation sequence of the full product. \textbf{b}, Margins of all 2,446 views on the deployment device (Apple M4, WebGPU, 16-bit), grouped by table. The smallest margin, 3.02 logits, belongs to the addition view 5 + 6 with no incoming carry. \textbf{c}, Exact answers on the pre-registered comparison of 220 problems (sums of 10\textendash{}1,000 digits and products of 2\textendash{}20-digit numbers, 20 per length; Methods), overall and on the longest sums and products. Sketchpad in rose, the strongest alternative in cobalt; Qwen3.5-0.8B is shown without chain of thought, its better setting. Per-length results: Extended Data Fig. \hyperref[fig-ed1]{1}.}\phantomsection\label{fig-sheet}
\end{figure}

Sketchpad makes this hypothesis hold by construction for a pretrained vision-language model. A runtime lays out a problem on a rendered sheet and shows the model one view at a time: three cells of the sheet. In one forward pass, the model reads the view through its vision encoder and outputs the step's results and the next operation. The runtime writes the outputs and moves the view, and nothing else passes between steps (Fig. \hyperref[fig-sheet]{1}a). The procedures are the classical pencil-and-paper ones, and a low-rank adapter\cite{hu2022lora} trained on their 2,446 views teaches them to Qwen3.5-0.8B.

We prove in Lean that exactness on the views is equivalent to exactness at all lengths, and that the restriction to views is necessary (Section \hyperref[sec-cert]{3}). We then certify the condition on the hardware that runs the model, a laptop GPU executing it in a web browser in 16-bit precision. All 2,446 views pass with a smallest margin of 3.02 logits, so the certificate tolerates any change of the logits smaller than 1.51. With these procedures the model adds, subtracts, multiplies and divides integers of any length and computes square roots and greatest common divisors. It answered all 220 problems of a pre-registered comparison in which the strongest of ten other models answered 217 (Section \hyperref[sec-calc]{4}). Because certified procedures compose, a symbolic runtime built on them solves, expands and factors polynomials with every number computed by the model (Section \hyperref[sec-algebra]{5}). Finally, a scalar gate on the adapter restores the base model exactly, and a dialogue adapter learns to call the sheet when a reply needs exact computation; it stays within 2 points of the base model on instruction following and MMLU but loses 5.8 points on fresh grade-school word problems (Section \hyperref[sec-gate]{6}). The guarantee covers the model's steps and is relative to the runtime, whose layouts, outer loops and reference procedures are conventional code; Section \hyperref[sec-cert]{3} states what they must satisfy.

\section{The sheet machine}\label{sec-machine}
Sketchpad has three parts: a sheet that holds all state, a model that maps a view of the sheet to the outputs of one step, and a runtime that moves the view and writes the outputs. The design serves one requirement: the input of each step must come from a finite set that can be listed in advance.

\textbf{Model.} Qwen3.5-0.8B\cite{qwen35} combines a vision encoder with a 24-layer language model in which six layers use full attention and eighteen use Gated DeltaNet\cite{yang2025gateddeltanet}. The only trained component is a rank-16 LoRA adapter on 150 projection matrices, with 10.2 million parameters (1.2\% of the model). We call it the arithmetic adapter. Section \hyperref[sec-gate]{6} derives from it a dialogue adapter, and preliminary adapters appear only as contrasts. Device measurements were made on the GPU of an Apple M4 laptop, which we call the M4.

\textbf{Views.} The sheet is a grid of 32-pixel cells, each blank or a digit, drawn from a fixed glyph set. A view is three cells, rendered as a strip of 32 \ensuremath{\times} 96 pixels that the vision encoder reads as three image tokens. Except in quotient estimation, the three cells form one column of the sheet. A step's prompt is an operation token, the view, \texttt{=\allowbreak{}} and two placeholder tokens, and the model's outputs are read at these three positions in one forward pass. The prompt contains nothing else. We call this the pinned-sheet policy; under it, the model's output is a function $f$ of the view alone.

\textbf{Tables.} For each operation, a table lists every view the operation can meet, with its correct outputs: column addition (\texttt{SUM}, 242 views), the five operations of long multiplication (1,553), subtraction (\texttt{SUB}, 242), estimation of a quotient digit (\texttt{EST}, 400) and decrement of an estimate (\texttt{DEC}, 9). With $\mathcal O$ the nine operations and $G$ the eleven glyphs, the rows form the set
\renewcommand{\theHequation}{eq-views}
\begin{equation}
V = \bigsqcup_{o \in \mathcal O} V_o, \qquad V_o \subseteq \{o\} \times G^3, \qquad |V| = 242 + 1{,}553 + 651 = 2{,}446,
\tag{1}\label{eq-views}
\end{equation}
whose terms count addition, multiplication and the calculator tables \texttt{SUB}, \texttt{EST} and \texttt{DEC} (Extended Data Table \hyperref[tbl-tables]{1}).

\textbf{Runtime.} For each procedure, a runtime of a few hundred lines lays out the operands, extracts views, calls the model and writes its outputs. It copies digits, as when a digit is brought down in division, but writes no digit of a result. In multiplication, the model's next-operation output determines the control flow. In division and square roots, the runtime fixes the outer loop and the model's outputs decide its branches. Independent procedures run in lockstep and share forward passes (Methods).

\textbf{Training.} We train the adapter directly on the tables, with cross-entropy, a hinge loss on the logit margin and a distillation penalty toward the base model on plain text\cite{li2017lwf} (Methods). Rows whose margin falls below 2 are replayed, and training stops when all rows have margins of at least 3. The tables are learned, not supplied by the base model. Without the adapter, Qwen3.5-0.8B passes 12 of the 2,446 views, all of them \texttt{SUM} views that copy a single digit; with eight solved views of the same table added to the prompt, it passes 158 of 2,380 (Methods).

\begin{figure}[tbp]
\centering
\includegraphics[width=\linewidth]{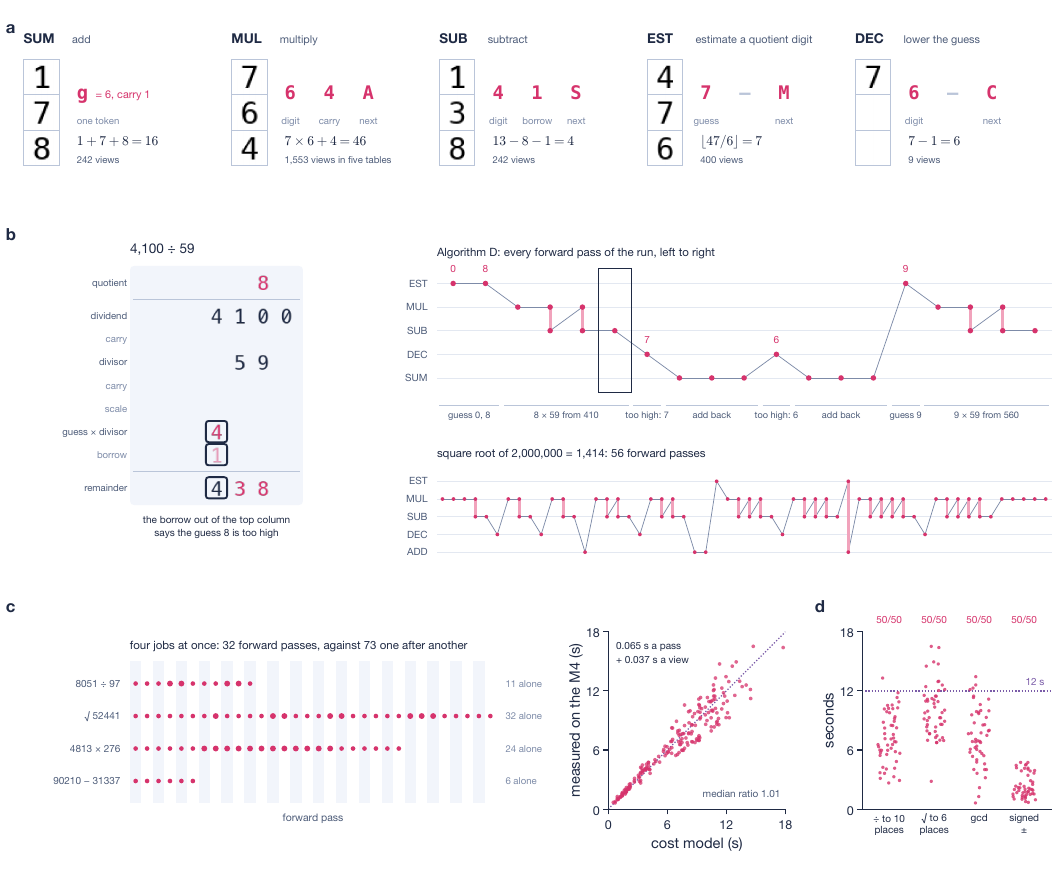}
\caption*{\textbf{Fig. 2 $|$} \textbf{The sheet machine.} \textbf{a}, One view and its outputs for each calculator table. \textbf{b}, Left: the division 4,100 \ensuremath{\div} 59; the borrow out of the leading column indicates that the estimate 8 is too high. Right: the operation sequence of the same division (Knuth's Algorithm D) and of the square root of 2,000,000 (56 forward passes). \textbf{c}, Batching four independent procedures: 32 forward passes instead of 73. Measured time against the cost model for the 200 problems of the browser test (median ratio 1.01). \textbf{d}, Time per problem in the browser test, all 200 exact.}\phantomsection\label{fig-machine}
\end{figure}
\section{Certification}\label{sec-cert}
Because each step is a function of its view, a check of the 2,446 views is a check of every run. This section makes the statement precise, shows that the restriction to views is necessary, and carries the check from exact arithmetic to the hardware.

At output position $k$ of a view $v$, let $z_k(v) \in \mathbb R^{\Sigma}$ be the model's logits over the vocabulary $\Sigma$ and $T_k(v)$ the correct token. The margin of $v$ is the smallest lead of a correct token over any other token,
\renewcommand{\theHequation}{eq-margin}
\begin{equation}
m(v) = \min_{k} \Big( z_k(v)_{T_k(v)} - \max_{a \neq T_k(v)} z_k(v)_a \Big),
\tag{2}\label{eq-margin}
\end{equation}
and $v$ passes if $m(v) > 0$. The model's output $f(v)$, the argmax at each position, then equals $T(v)$ under any tie rule. A certificate records $m(v)$ for all $v \in V$.

Let $C$ be the configurations of a procedure (sheet contents, view position and control state), $\mathcal V \supseteq V$ all views and $A$ the outputs. At configuration $c_t$, the runtime either halts, $\mathrm{view}(c_t) = \bot$, or presents the view $\mathrm{view}(c_t) \in \mathcal V$ and writes the outputs:
\renewcommand{\theHequation}{eq-step}
\begin{equation}
c_{t+1} = \mathrm{paint}\big(c_t,\, f(\mathrm{view}(c_t))\big).
\tag{3}\label{eq-step}
\end{equation}
The reference run uses the table $T\colon \mathcal V \to A$ in place of $f$.

\begin{thmbox}\phantomsection\label{thm-p}\noindent\textbf{Theorem 1} (Exactness on views).\enspace\itshape Let $R_T(I)$ be the set of views presented on reference runs from a set $I$ of initial configurations, and assume that $\mathrm{paint}(c, \cdot)$ is injective whenever $\mathrm{view}(c) \neq \bot$. The model's run from every $c_0 \in I$ equals the reference run at every step if and only if $f(v) = T(v)$ for all $v \in R_T(I)$. In particular, if $R_T(I) \subseteq V$ and $m(v) > 0$ for all $v \in V$, every run of every length is exact; this direction does not need injectivity.\upshape
\end{thmbox}
\begin{proofbox}\noindent\textit{Proof sketch.}\enspace If $f = T$ on $R_T(I)$ and the runs agree at $c_t$, then $\mathrm{view}(c_t) \in R_T(I) \cup \{\bot\}$ and (\hyperref[eq-step]{3}) gives the same $c_{t+1}$; induction on $t$ completes sufficiency. Conversely, if the runs agree and $v = \mathrm{view}(c)$ for a configuration $c$ of a reference run, then $\mathrm{paint}(c, f(v)) = \mathrm{paint}(c, T(v))$, and injectivity gives $f(v) = T(v)$.
\end{proofbox}
Theorem \hyperref[thm-p]{1} separates what the network must get right from what the runtime must get right. The network's part, $f = T$ on $V$, is finite, and we certify it below. The runtime's part is that reference runs present only table views, $R_T(I) \subseteq V$, and compute the intended functions. The first follows from the layouts and normalisations of Section \hyperref[sec-calc]{4}, and the runtime asserts it at each step; the second is the correctness of the classical procedures. Neither is formalised in Lean (Methods).

The restriction to views is necessary. If a step's context also contains earlier steps, as in chain of thought, no finite test suffices.

\begin{thmbox}\phantomsection\label{thm-nofinite}\noindent\textbf{Theorem 2} (Necessity of the pinned-sheet policy).\enspace\itshape Suppose instead that each step's context retains the earlier steps. There is a two-state table and a family of decoders, each a correct lookup plus one attention head, such that for every finite set of test contexts some member passes every row of the table, and every test context that begins with the prompt and ends in a row, with positive margins, yet errs on a longer run. Under the pinned-sheet policy the same member is exact at every length.\upshape
\end{thmbox}
\begin{proofbox}\noindent\textit{Proof sketch.}\enspace On a context with $n$ tokens after a sink token, $k$ of them events, the head writes $4Rk/(e^{c} + n)$ to the logit of the wrong state, where $R$ is the lookup's margin and $c$ the head's score on the sink. With $e^{c} = 3M + 5$, for $M$ the length of the longest test context, the write exceeds $R$ only on longer contexts (Appendix \hyperref[app-proofs]{A}). This is an instance of Moore's result that no finite experiment identifies a machine with an unbounded number of states\cite{moore1956gedanken}.
\end{proofbox}
The theorems assume exact arithmetic, but the model runs in floating point, and the logits on the deployment path differ from those of any reference computation. The differences can be large: a 4-bit quantised decoder moved some margins of a preliminary adapter from +9.4 to \ensuremath{-}8.9. Margins make small differences harmless.

\begin{figure}[tbp]
\centering
\includegraphics[width=\linewidth]{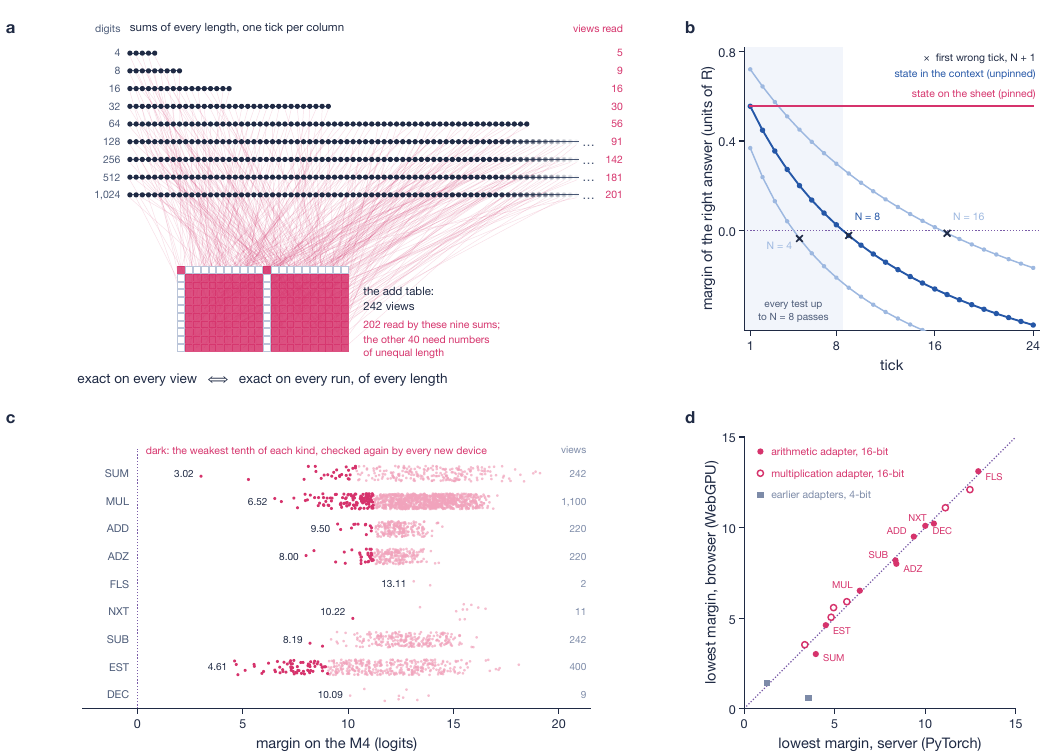}
\caption*{\textbf{Fig. 3 $|$} \textbf{Certification.} \textbf{a}, Sums of increasing length read only views from the 242-view addition table; nine sums of 4 to 1,024 digits reach 202 of them. \textbf{b}, The counterexample of Theorem \hyperref[thm-nofinite]{2}: with state in the context, a decoder passes all tests up to $N$ steps and fails at step $N+1$ ($N$ = 4, 8, 16; the sink score satisfies $e^{c} = 2N + 3$, Appendix \hyperref[app-proofs]{A}); under the pinned-sheet policy the margin is constant. \textbf{c}, Margins of all 2,446 views on the M4, by operation; dark points mark the lowest tenth, which is rechecked on each new device. \textbf{d}, Smallest margin per operation on the server (PyTorch, 32-bit) against the browser (WebGPU): 16-bit decoders lie on the identity line; 4-bit decoders of preliminary adapters deviate by up to 3 logits.}\phantomsection\label{fig-cert}
\end{figure}

\begin{thmbox}\phantomsection\label{thm-transfer}\noindent\textbf{Theorem 3} (Robustness to perturbation).\enspace\itshape Let $z, \tilde z \in \mathbb R^{O}$ over a finite set $O$, and $y \in O$. If $\lVert z - \tilde z \rVert_\infty \le \tau$ and $z_y - \max_{a \neq y} z_a > 2\tau$, then $y$ is the unique maximiser of $\tilde z$. Hence, if $R_T(I) \subseteq V$, $m(v) > 2\tau$ for all $v \in V$, and at every step the logits on the execution path for each view $v \in V$ are within $\tau$ of $z_k(v)$ at every output position, every run from $I$ is exact.\upshape
\end{thmbox}
\begin{proofbox}\noindent\textit{Proof sketch.}\enspace If $a \neq y$ maximised $\tilde z$, then $\tilde z_a \le z_a + \tau < z_y - \tau \le \tilde z_y$. In a run, induction on $t$ as for Theorem \hyperref[thm-p]{1}: while the run agrees with the reference run, its view lies in $V$ and each output is the certified one. A cached-prefix implementation of the pinned step computes the logits of the full prompt exactly (Lemma \hyperref[lem-cache]{4}), so it is a path with $\tau = 0$ in exact arithmetic.
\end{proofbox}
We therefore certify on the deployment path itself. In 32-bit PyTorch, the smallest margin over the 2,446 views is 3.96. On the deployment path, ONNX Runtime Web on WebGPU\cite{onnxruntime,transformersjs,webgpu} in 16-bit precision, all views pass, with smallest margins of 3.02 for addition, 6.52 for multiplication and 4.61 for the calculator tables (Fig. \hyperref[fig-cert]{3}c). Per table, these differ from the server's smallest margins by less than one logit, whereas the 4-bit decoders of preliminary adapters deviate by up to 3 (Fig. \hyperref[fig-cert]{3}d). The full check takes 3.5 minutes on the M4. By Theorem \hyperref[thm-transfer]{3}, a margin of 3.02 tolerates any change of the logits smaller than 1.51. The certificate scores one view per forward pass, whereas inference batches up to 32 views, and batching changes the 16-bit logits. With every view placed in random batches of 2, 4 and 32 for the arithmetic adapter, and of 4 and 32 for the dialogue adapter of Section \hyperref[sec-gate]{6}, no logit moved by more than 1.09 and no margin by more than 0.64. The batched paths tested therefore satisfy Theorem \hyperref[thm-transfer]{3} with $\tau = 1.09$, below half the smallest margin of 3.02. On a different GPU, the client recomputes the certificate on request; by default it recomputes the lowest-margin tenth of each table. Appendix \hyperref[app-proofs]{A} gives complete proofs and the corresponding Lean 4 theorems.

\section{Arithmetic}\label{sec-calc}
Theorem \hyperref[thm-p]{1} applies to a procedure only if every view it reaches is a table row. Signed addition, subtraction and schoolbook multiplication meet this by their layout, since each column holds two digits and a carry or borrow; the borrow out of the final column of a subtraction determines the sign. Division needs more, because a quotient digit is estimated from leading digits alone. Following Knuth's Algorithm D\cite{knuth1997taocp2}, both operands are first multiplied on the sheet by $\lfloor 10/(d_0 + 1) \rfloor$, where $d_0$ is the divisor's leading digit. The quotient is unchanged, and the scaled divisor $b$ has a leading digit $d$ of at least 5. For a partial remainder $r < 10b$ with leading digits $r_1 r_2$, \texttt{EST} returns
\renewcommand{\theHequation}{eq-est}
\begin{equation}
\hat q = \min\Big(9, \Big\lfloor \frac{10 r_1 + r_2}{d} \Big\rfloor\Big), \qquad q \le \hat q \le q + 2,
\tag{4}\label{eq-est}
\end{equation}
where $q = \lfloor r/b \rfloor$ is the true digit; the bounds are Knuth's Theorems A and B for base 10 and $d \ge 5$, and the inputs range over the 400 rows of the table. A borrow out of the multiply-subtract pass $r - \hat q\, b$ shows that the estimate is too large: \texttt{DEC} lowers it and the divisor is added back, at most twice.

Square roots use the digit-by-digit method on $c^2 a$, with $c \in \{1, 2, 4, 5\}$ chosen so that the root, if it has more than one digit, begins with 25 to 49. Twice the partial root then begins with a digit of at least 5, and the same estimation table serves every later digit (Methods). All scalings run on the sheet, and $\lfloor \sqrt a \rfloor = \lfloor \lfloor \sqrt{c^2 a} \rfloor / c \rfloor$. Greatest common divisors use Euclid's algorithm.

\textbf{Browser test.} We pre-registered 200 problems: 50 each of division to 10 decimal places, square roots to 6 places, greatest common divisors of integers up to 999, and signed sums and differences of up to 20 digits. Run one at a time on the M4, all 200 answers were exact. Median times were 7.0 s for division, 9.8 s for square roots, 7.0 s for greatest common divisors and 2.0 s for sums. Of the 200 problems, 184 met the pre-registered target of 12 s, and the slowest took 16.5 s (Fig. \hyperref[fig-machine]{2}d; Extended Data Table \hyperref[tbl-t3]{2}). Measured times follow a cost model linear in forward passes and views, with a median ratio of 1.01 (Fig. \hyperref[fig-machine]{2}c; Methods).

\begin{figure}[tbp]
\centering
\includegraphics[width=\linewidth]{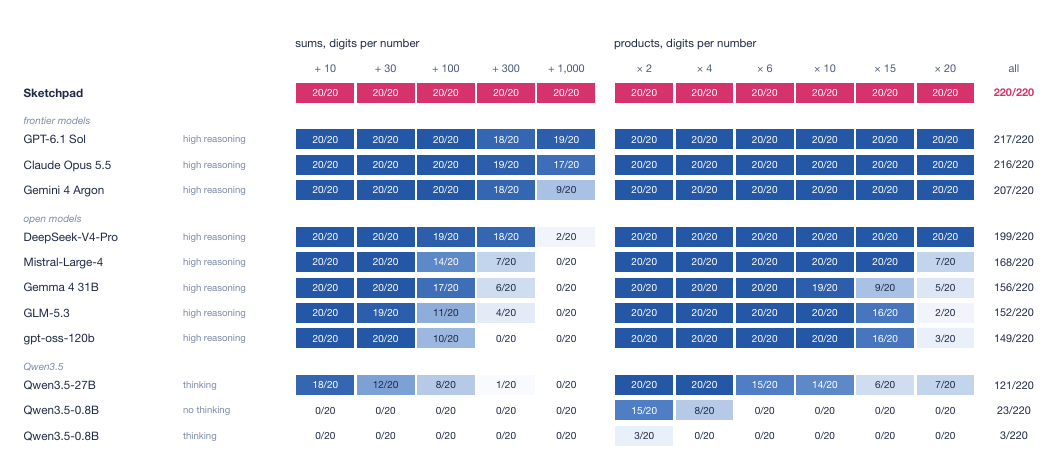}
\caption*{\textbf{Extended Data Fig. 1 $|$} \textbf{Exact answers by length.} Problems answered exactly, of 20 per length, for every system in the comparison: sums of 10 to 1,000 digits (left) and products of 2- to 20-digit numbers (right).}\phantomsection\label{fig-ed1}
\end{figure}

\textbf{Comparison.} We compared Sketchpad with ten language models on 220 pre-registered problems: sums of 10 to 1,000 digits and products of 2- to 20-digit numbers, 20 per length, with the same prompt and a budget of 32,768 output tokens (Methods). Sketchpad answered all 220. GPT-6.1 Sol, Claude Opus 5.5 and Gemini 4 Argon answered 217, 216 and 207, exact on all products, with all of their errors on sums of 300 or 1,000 digits. Five open-weight models answered between 149 and 199, failing mostly by exhausting the token budget on long sums and long products. Qwen3.5-27B answered 121, and Qwen3.5-0.8B, the base of Sketchpad, 23 without chain of thought and 3 with it (Fig. \hyperref[fig-sheet]{1}c; Extended Data Fig. \hyperref[fig-ed1]{1} and Table \hyperref[tbl-baselines]{5}).

\section{Algebra}\label{sec-algebra}
An algebraic derivation is a sequence of calls to certified procedures, whose inputs are transcribed from the problem and from earlier results. Its correctness then reduces to three conditions.

\begin{thmbox}\phantomsection\label{thm-compose}\noindent\textbf{Theorem 4} (Composition).\enspace\itshape Consider $K$ calls with results $r = (r_1, \dots, r_K)$, in which the input $x_i = q_i(r_{<i})$ of call $i$ depends only on the results $r_{<i} = (r_1, \dots, r_{i-1})$ of earlier calls and the answer $g(r)$ only on $r$; let $q^*_i$, $T^*_i$ and $g^*$ be the intended inputs, functions and combining program, with results $r^*_i = T^*_i(q^*_i(r^*_{<i}))$. If every call returns its table's output on the input it received, $r_i = T_i(x_i)$, as Theorem \hyperref[thm-p]{1} guarantees for a certified procedure, then a wrong answer requires, for some $i \le K$, a wrong table, $T_i(q^*_i(r^*_{<i})) \neq T^*_i(q^*_i(r^*_{<i}))$, or a wrong transcription, $q_i(r^*_{<i}) \neq q^*_i(r^*_{<i})$, or else a wrong combining program, $g(r^*) \neq g^*(r^*)$, and for any distribution of transcripts
\renewcommand{\theHequation}{eq-compose}
\begin{equation}
\Pr\big[g(r) \neq g^*(r^*)\big] \le \Pr[\textup{some table wrong}] + \Pr[\textup{some transcription wrong}] + \Pr\big[g(r^*) \neq g^*(r^*)\big].
\tag{5}\label{eq-compose}
\end{equation}\upshape
\end{thmbox}
The certificate excludes the first failure. The other two, transcribing a problem into calls and combining their results, belong to the symbolic runtime, which is conventional code, and we guard them by checking each answer before it is returned.

\begin{figure}[tbp]
\centering
\includegraphics[width=\linewidth]{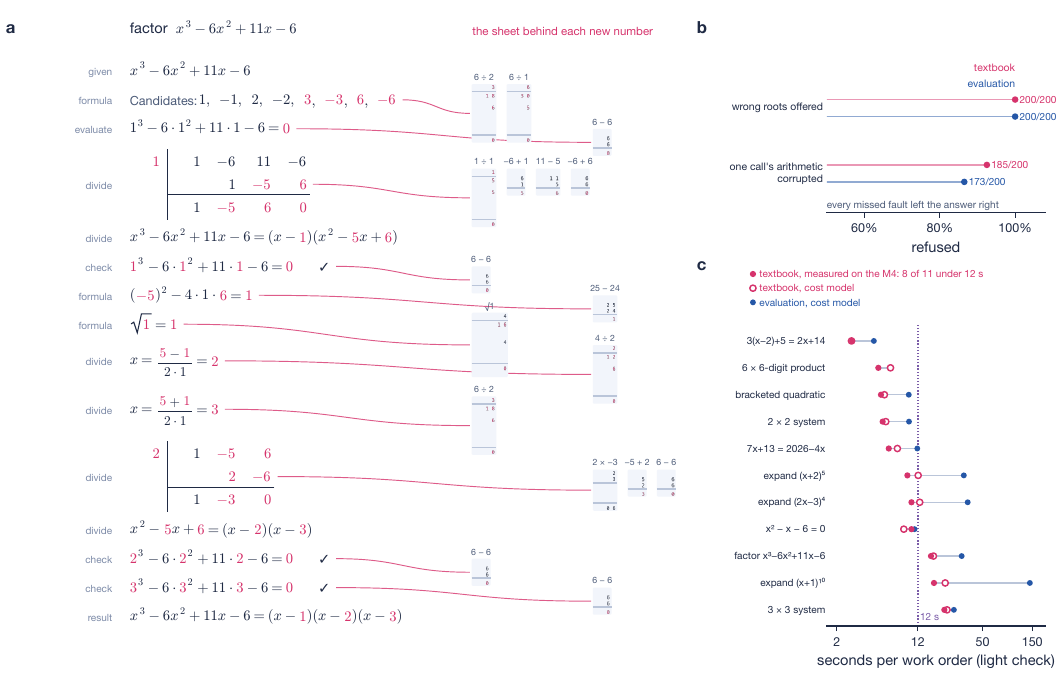}
\caption*{\textbf{Fig. 4 $|$} \textbf{Algebra.} \textbf{a}, Factorisation of $x^3-6x^2+11x-6$. Numbers in rose were computed by the model; each is linked to the column operation that produced it. \textbf{b}, Rejection rates under fault injection: incorrect roots (200 of 200 for both routes) and one corrupted arithmetic result (185 of 200 for the textbook route, 173 of 200 for evaluation). \textbf{c}, Time per benchmark problem: textbook route measured on the M4 (rose) and under the cost model (open rose); evaluation route under the cost model (cobalt).}\phantomsection\label{fig-algebra}
\end{figure}

The runtime follows the textbook procedures. It expands products and collects terms, solves linear equations by rearrangement and quadratics by the quadratic formula, expands powers by the binomial theorem with coefficients from Pascal's rule, factors by the rational root theorem and synthetic division, and solves linear systems by elimination (Fig. \hyperref[fig-algebra]{4}a). Irrational roots are returned as radicals with square factors removed, for example $3 \pm \sqrt 5$. Rational roots are checked by substitution and irrational pairs by Vieta's relations in integer arithmetic, both exactly. Expansions are evaluated at $x = 1$, which detects any single wrong coefficient but not errors that cancel there (Methods). Under fault injection, the checks rejected 200 of 200 incorrect roots and 185 of 200 runs with one corrupted arithmetic result; in the other 15 runs the corruption did not change the answer (Fig. \hyperref[fig-algebra]{4}b).

Computer algebra reduces algebra to arithmetic in another way, by evaluation at points followed by interpolation\cite{kaltofen1990blackbox} or Kronecker substitution\cite{kronecker1882}. Under the cost model, the textbook route is faster on ten of eleven benchmark problems and equal on the eleventh, a plain multiplication: 2.8 s against 4.5 s for a linear equation, 12.1 s against 33.1 s for $(x+2)^5$ and 16.9 s against 31.5 s for factoring $x^3 - 6x^2 + 11x - 6$. Measured on the M4, eight of the eleven finish within the 12-s target; $(x+1)^{10}$, a cubic factorisation and a 3 \ensuremath{\times} 3 system take 16 to 22 s (Fig. \hyperref[fig-algebra]{4}c; Extended Data Table \hyperref[tbl-algebra]{3}).

\section{Gating}\label{sec-gate}
\begin{figure}[tbp]
\centering
\includegraphics[width=\linewidth]{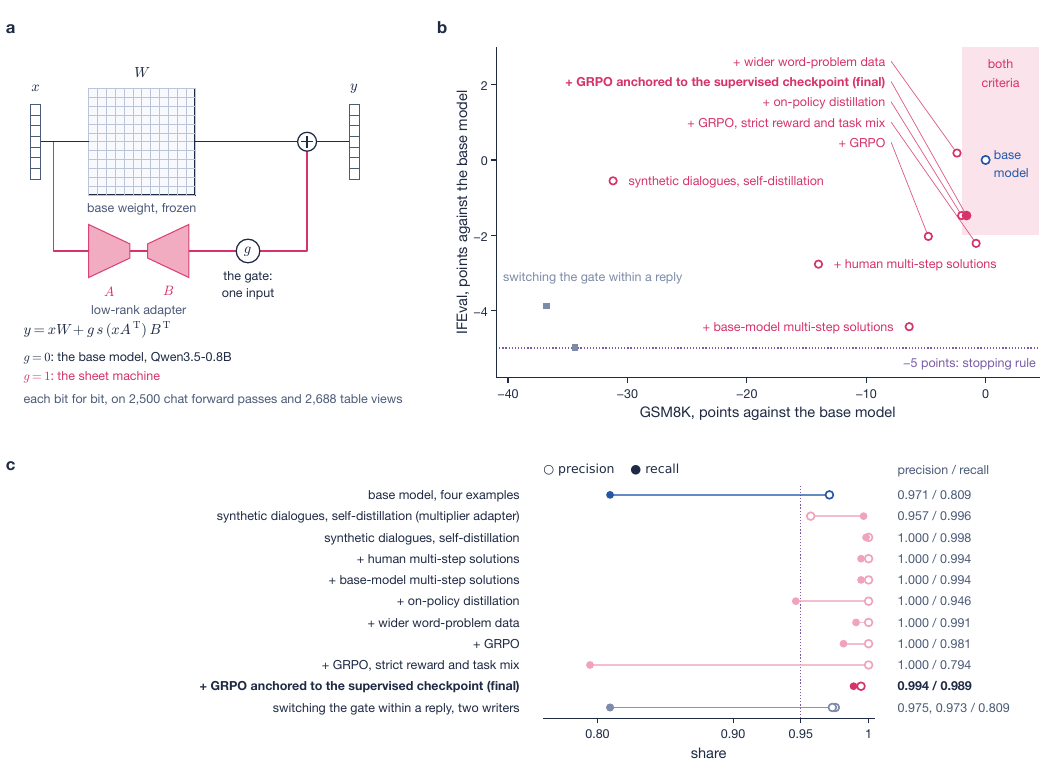}
\caption*{\textbf{Fig. 5 $|$} \textbf{Gating.} \textbf{a}, A scalar input $g$ multiplies every adapter branch; at $g=0$ the model is identical to the base model, at $g=1$ it executes the sheet procedures. \textbf{b}, IFEval against GSM8K, in points relative to the base model on the development items, for the dialogue fine-tunes and for switching the gate within a reply (on fresh items the dialogue adapter's GSM8K difference is \ensuremath{-}5.8); the shaded region satisfies both criteria, and the dotted line at \ensuremath{-}5 points is the pre-registered stopping rule. \textbf{c}, Precision and recall of invoking the sheet on held-out problems: base model with four examples (cobalt) and fine-tunes (rose).}\phantomsection\label{fig-gate}
\end{figure}

Fine-tuning for arithmetic damages other abilities. A preliminary arithmetic adapter changed perplexity on plain text by only 0.65\%, yet it lowered strict instruction-following accuracy on IFEval\cite{zhou2023ifeval} by 16.4 points, an instance of catastrophic forgetting\cite{mccloskey1989catastrophic,kirkpatrick2017ewc}. We therefore multiply every adapter branch by a scalar graph input $g$:
\renewcommand{\theHequation}{eq-gate}
\begin{equation}
y = xW + g\, s\, (xA^{\top})B^{\top}, \qquad y\big|_{g=0} = xW,
\tag{6}\label{eq-gate}
\end{equation}
where $W$ is the frozen weight, $A$ and $B$ are the adapter matrices and $s$ is the LoRA scale (Fig. \hyperref[fig-gate]{5}a). At $g = 0$ each branch contributes an exact floating-point zero, provided it is finite, so the gated model is the base model; we confirmed this bit for bit for logits, hidden states and cached states (Methods). At $g = 1$, the certificate on the M4 reproduces all 2,446 margins.

The gate is set by the runtime, not by the model. Switching it within a reply, so that the base model writes the prose and the adapter only the sheet's program, leaves the decision to compute with the base model, whose recall is 0.809 (Methods). We therefore keep $g = 1$ throughout a conversation and trained the dialogue adapter, starting from the arithmetic adapter, to decide by itself: it emits a short program for the sheet when a reply requires exact computation; generation then pauses, and the runtime executes the program and returns its results into the reply (Methods). Training combines supervised data with self-distillation\cite{li2017lwf} and on-policy distillation\cite{agarwal2024gkd,lu2025onpolicy} toward the gated-off model, which limits forgetting\cite{shenfeld2025razor}. Reinforcement learning with GRPO\cite{shao2024grpo} follows, against an execution-based reward and with a KL penalty toward the supervised checkpoint, and every stage keeps the table objective.

Against criteria fixed before training (Methods), the dialogue adapter invokes the sheet on held-out phrasings with precision 0.994 and recall 0.989, against 0.971 and 0.809 for the base model prompted with four examples. It writes the correct computation for 92.8\% of word problems, 96.7\% of equations and 98.3\% of expressions embedded in prose. On the benchmark items used during development, it stayed within 1.6 points of the base model on IFEval (\ensuremath{-}1.5 \ensuremath{\pm} 2.0), GSM8K\cite{cobbe2021gsm8k} (\ensuremath{-}1.6 \ensuremath{\pm} 3.4) and MMLU\cite{hendrycks2021mmlu} (\ensuremath{-}1.4 \ensuremath{\pm} 1.6), each less than one standard error from the criterion of \ensuremath{-}2. Because it was the fifth candidate scored on those items, we rescored it on items not used before, against predictions fixed in advance. MMLU held, at \ensuremath{-}2.0 \ensuremath{\pm} 1.1 on 1,000 questions, but GSM8K did not: \ensuremath{-}5.8 \ensuremath{\pm} 1.7 on the 1,069 remaining test problems, so the retention criterion is not confirmed for multi-step word problems. The loss appears both where the adapter calls the sheet and where it answers directly, which places it in multi-step reasoning rather than in the sheet. The same weights at $g = 1$ execute the arithmetic procedures: on the M4, all 2,446 views pass, with smallest margins of 6.67 for addition, 6.53 for multiplication and 4.56 for the calculator tables. Across recipes, IFEval and GSM8K trade off against each other (Fig. \hyperref[fig-gate]{5}b). On-policy distillation removes most of the trade-off, and reinforcement learning reintroduces it unless anchored to the supervised checkpoint (Extended Data Table \hyperref[tbl-chat]{4}).

\section{Mechanism}\label{sec-inside}
\begin{figure}[tbp]
\centering
\includegraphics[width=\linewidth]{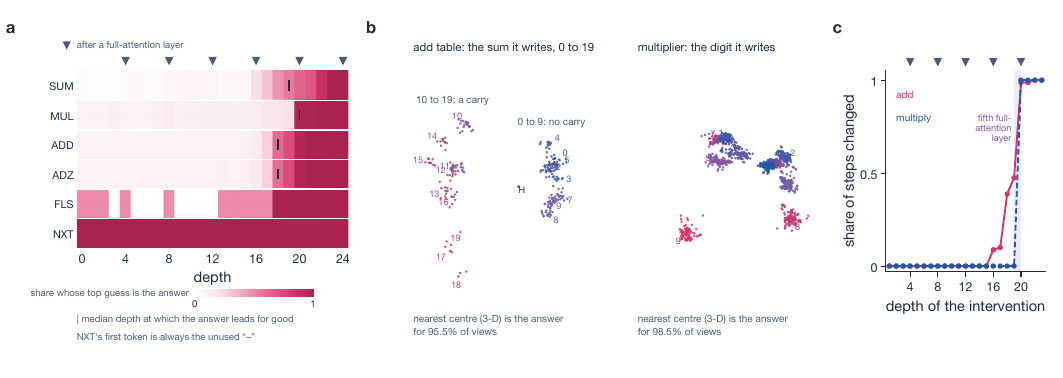}
\caption*{\textbf{Fig. 6 $|$} \textbf{Mechanism.} Arithmetic adapter. \textbf{a}, Logit lens: fraction of views whose top token at each depth equals the final output, by operation; ticks mark the median decision depth, triangles the ends of full-attention layers. \textbf{b}, Final-layer states projected onto three discriminant directions, coloured by output, for addition (0\textendash{}19) and multiplication digits; nearest-centre shares are measured in this projection (Methods). \textbf{c}, Fraction of 80 steps whose output a class-mean intervention changes to its target, against the depth at which it is applied, for addition and multiplication; the shaded band is the fifth full-attention layer.}\phantomsection\label{fig-inside}
\end{figure}

The certificate establishes what the model outputs on each view, not where in the network the output is decided. We applied the logit lens\cite{nostalgebraist2020logitlens,belrose2023tunedlens} to the arithmetic adapter at the first output position: with $h_\ell$ the residual state after depth $\ell$, and $W_U$ and $\mathrm{norm}$ the final unembedding and normalisation,
\renewcommand{\theHequation}{eq-lens}
\begin{equation}
\hat p_\ell = \operatorname{softmax}\big(W_U\, \mathrm{norm}(h_\ell)\big).
\tag{7}\label{eq-lens}
\end{equation}
The correct output becomes and remains the top token at a median depth of 19 for addition, 20 for multiplication and 18 for the addition steps of the multiplier; up to depth 15 it is the top token for at most 13\% of views in any of these tables (Fig. \hyperref[fig-inside]{6}a). At the final layer, states cluster by output, and sums with a carry (10\textendash{}19) separate from those without (Fig. \hyperref[fig-inside]{6}b).

To test whether this structure is causal, a class-mean intervention\cite{turner2023actadd,rimsky2024caa} moves the state of a view with output $a$ toward a target output $a'$,
\renewcommand{\theHequation}{eq-steer}
\begin{equation}
h_\ell \leftarrow h_\ell + \mu^{\ell}_{a'} - \mu^{\ell}_{a},
\tag{8}\label{eq-steer}
\end{equation}
where $\mu^{\ell}_a$ is the mean of $h_\ell$ over the table's views with output $a$. Applied after depth 20, the end of the fifth full-attention layer, it changes the output to $a'$ in 79 of 80 addition steps and 80 of 80 multiplication steps; applied at any depth up to 15, it changes none (Fig. \hyperref[fig-inside]{6}c). For multiplication the transition is abrupt: no step changes at depth 19 and every step at depth 20, so the digit is decided in the fifth full-attention layer. For addition it is gradual, with 31 and 38 of 80 steps changed at depths 18 and 19, and the same layer completes a decision that the preceding layers have begun, consistent with the median decision depth of 19 under the logit lens.

\section{Discussion}\label{sec-discussion}
A pretrained vision-language model of 0.8 billion parameters, restricted to a bounded view of its working memory, can be certified exact on inputs of every length by checking 2,446 views on the device that runs it. On the longest sums of our comparison, where the strongest frontier models made silent errors, its answers carry this guarantee, and the same certified procedures support symbolic algebra and sit behind an exact gate in a dialogue model.

The certificate covers the model's steps. The runtime that sequences them, the transcription of a problem into calls and the translation of a word problem into a computation are conventional code or uncertified model behaviour; the dialogue adapter mistranslates about 7\% of held-out word problems, and on fresh grade-school problems it reasons less well than the base model, 5.8 points below it on GSM8K. In algebra, the answer checks guard these parts without being complete: they verify the roots found, not that all roots were found, and evaluation at a point is a spot check. The guarantee holds for the execution path on which the margins were measured, and on a new GPU it requires the full check, of which the client by default repeats only the lowest-margin tenth. Inputs must be rendered in the model's glyph set, the algebra is limited to polynomials of low degree in one variable and linear systems of up to three equations, and a step takes about 0.1 s on a laptop GPU.

Theorem \hyperref[thm-nofinite]{2} bears on methods that estimate accuracy on samples. When a step reads the steps before it, as in chain of thought and in most formats studied for length generalisation, no finite test certifies the model, however long the tested inputs. Context disciplines such as the inductive scratchpad\cite{abbe2024globality} and PENCIL\cite{yang2025pencil} bound what a step reads in order to generalise. The pinned-sheet policy, like the decoding of Lewandowski et al.\cite{lewandowski2026universal}, is the limit in which a step reads only a bounded view, and Theorem \hyperref[thm-p]{1} shows that in this limit a finite check suffices. The price is a runtime that sequences the steps, and the gain is a guarantee that does not depend on which inputs were tested.

\section*{Methods}\label{methods}
\subsection*{Model and adapter}
The language model of Qwen3.5-0.8B\cite{qwen35} has 24 layers with hidden size 1,024; layers 4, 8, \ldots{}, 24 (numbered from 1) use full attention (8 query and 2 key-value heads), the others Gated DeltaNet\cite{yang2025gateddeltanet}. The LoRA adapter (rank 16, scale 2) is applied to the query, key, value and output projections of the 6 attention layers (24 matrices), the two input projections and the output projection of the 18 Gated DeltaNet layers (54) and the three projections of each of the 24 feed-forward blocks (72): 150 matrices and 10,223,616 parameters, against 873,438,784 in the base model. For deployment the model is exported to ONNX, with a 16-bit vision encoder and language model and 4-bit token embeddings (16-bit, looked up inside the decoder graph, for the dialogue adapter); the adapter is folded into the decoder graph. Two additional graph inputs implement the gate and the steering intervention, and one additional output returns hidden states.

\subsection*{Views, prompts and tables}
Glyphs are fixed 32 \ensuremath{\times} 32 sprites of blank and the ten digits, so the set of possible views is finite and each view is an exact image. The prompt of a step is its operation token, the 32 \ensuremath{\times} 96 image, \texttt{=\allowbreak{}} and two placeholder tokens, padded with further placeholders to a common length. Since the model is causal, padding after the output positions does not affect the outputs (Lemma \hyperref[lem-cache]{4}), so steps of different operations can share a batch. The tables are exhaustive over each operation's views; for example, \texttt{MUL} covers every multiplicand digit or blank, every multiplier digit and every carry (blank, 0\textendash{}8; 11 \ensuremath{\times} 10 \ensuremath{\times} 10 = 1,100 views). The view of \texttt{EST} is assembled from three cells that do not share a column: the leading remainder digit $r_1$, the next remainder digit $r_2$ and the divisor's leading digit $d$, in the column of $r_2$. A blank $r_1$ is written as 0 before the view is read, so \texttt{EST} covers $r_1 \in \{0, \dots, d\}$, $r_2 \in \{0, \dots, 9\}$ and $d \in \{5, \dots, 9\}$, $10 \sum_{d=5}^{9} (d + 1) = 400$ views; \texttt{DEC} reads the guess and two blank cells. That the procedures reach no other view follows from the layout and, for division and square roots, from normalisation; the runtime checks it at every step, and a reference implementation checked it on 1,000 random problems per procedure. No view outside the tables occurred in any experiment.

\subsection*{Training}
Each training step draws a batch $R$ of rows from every table, together with rows replayed from the last certificate. With $z_k(v;\theta)$ the logits at output position $k$ of row $v$, $K_v \in \{1, 3\}$ the number of output positions and $m_k(v;\theta)$ the margin at position $k$, the term inside the minimum in (\hyperref[eq-margin]{2}), the loss is
\renewcommand{\theHequation}{eq-loss}
\begin{equation}
\begin{aligned}
\mathcal L(\theta) = {} & \frac{1}{|R|} \sum_{v \in R} \frac{1}{K_v} \sum_{k=1}^{K_v} \Big[ -\log \operatorname{softmax}\big(z_k(v;\theta)\big)_{T_k(v)} + \max\big(0,\, h - m_k(v;\theta)\big) \Big] \\
& + \frac{\lambda}{n} \sum_{j=1}^{n} \mathrm{KL}\big(p_0(\cdot \mid x_{\le j}) \,\big\|\, p_\theta(\cdot \mid x_{\le j})\big),
\end{aligned}
\tag{9}\label{eq-loss}
\end{equation}
with hinge target $h = 4$ and $\lambda = 1$, where $x$ is a passage of $n = 128$ tokens of plain text and $p_0$ the base model\cite{li2017lwf}. Every 100 steps the full certificate is computed and rows with margins below $h/2 = 2$ are replayed until the next; training stops when every row has a margin of at least 3. The arithmetic adapter was initialised from a preliminary adapter certified on the addition and multiplication tables, with the subtraction, estimation and decrement tables added. It passed every row at step 1,200 and was stopped at step 1,600 (19.8 minutes on one H100 GPU; learning rate on a cosine schedule from $2\times10^{-4}$ to $2\times10^{-5}$ over 4,000 steps). Its plain-text perplexity is 1.0048 times that of the base model.

\subsection*{Certification}
Margins are computed over the full vocabulary of 248,320 tokens. The server certificate uses PyTorch in 32-bit floating point. The device certificate runs the deployed files (transformers.js 4.2.0\cite{transformersjs}, ONNX Runtime Web on WebGPU) through the inference code path and batched prompt form, and records every margin with the device, library version, file hashes and glyph set (Appendix \hyperref[app-cert]{B}). Padding the addition prompt to the batched form changed no margin by more than 0.0018. Two device certificates, with and without the gate, agree on all margins to $10^{-3}$. The certificate is bound to the model files by their hashes (Appendix \hyperref[app-cert]{B}). The batch check scores each of the 2,446 views alone and inside a batch of randomly chosen other views, in the padded prompt form of inference, and records the largest change of any logit over the full vocabulary and of the view's margin. For each batch size, one random partition covers all views once: sizes 2, 4 and 32 for the arithmetic adapter, and 4 and 32 for the dialogue adapter. The check samples batch compositions; it does not bound all of them.

The base-model control scores the same 2,446 views with the base model, in 32-bit PyTorch and the prompt form of the server certificate. Zero-shot, it passes 12 views, all of them \texttt{SUM} views, and 1.1\% of output positions are correct. With eight solved views of the same table (fixed seed), each followed by its outputs, prefixed to every query, it passes 35 of the 2,380 views not among the examples, with outputs read at the placeholder positions, and 158 when each output after the first is read after the correct earlier outputs, the form the examples show (45.8\% of positions); no table passes completely in any setting.

\subsection*{Procedures and timing}
Independent procedures, such as the products of an algebra step, run in lockstep, with one view from each in a shared batch. On the M4, a computation of $n_{\text{pass}}$ forward passes over $n_{\text{view}}$ views takes
\renewcommand{\theHequation}{eq-cost}
\begin{equation}
t \approx \alpha\, n_{\text{pass}} + \beta\, n_{\text{view}}, \qquad \alpha = 0.065\ \text{s}, \quad \beta = 0.037\ \text{s},
\tag{10}\label{eq-cost}
\end{equation}
so batching reduces $n_{\text{pass}}$ and leaves $n_{\text{view}}$ unchanged.

Subtraction ends on a row that reports a final borrow, in which case the operands are swapped and the sign recorded. In division and square roots, a multiply-subtract pass of the divisor (or trial divisor) by a guess runs \texttt{MUL} one column ahead of \texttt{SUB}, two views per forward pass, with separate rows for the carry and the borrow. The runtime checks each next-operation output against the pass (\texttt{A} within it, \texttt{Z} at its end for \texttt{MUL}, \texttt{S} for \texttt{SUB}), and the borrow \texttt{SUB} writes out of the top column is the branch: no borrow keeps the guess, a borrow lowers it with \texttt{DEC}. In division the divisor is then added back with \texttt{SUM}, and the carry out of the top column, which cancels the borrow, decides whether to lower again; in square roots the pass is repeated from the saved remainder. A guess of 0 from \texttt{EST} comes with the next-operation \texttt{N} and skips the pass. In square roots, the first root digit is at most 4, and when it is 2 the second is at least 5; the runtime starts these at 4 and 9, the largest values the scale allows, so that the first is lowered at most twice and the second at most four times, and every other digit at the estimate of \texttt{EST}. Each root digit is thus fixed by the model's multiply-subtract and \texttt{DEC} steps; the runtime copies a lowered digit into the trial divisor, and a carry out of the trial divisor's last digit is added to the next column with \texttt{ADD}, whose running-total cell is then blank. Greatest common divisors repeat division with remainder until the remainder the model writes is zero. In square roots, the trial divisor for the next digit, twice the partial root, is formed on the sheet by adding the last digit to the previous trial divisor with \texttt{ADD}.

For square roots, write $a = A \cdot 100^k$ with $1 \le A < 100$. The runtime takes $c = 4$ if $A < 1.5625$, $c = 2$ if $A < 6.25$, $c = 1$ if $A < 25$ and $c = 5$ otherwise, comparing the first six digits of $a$, aligned to pairs, with 15625, 62500 and 250000. Then $c \sqrt A$ lies in $[2.5, 5)$ if $A < 25$ and in $[25, 50)$ otherwise, so the root of $c^2 a$ begins with 25 to 49 when it has more than one digit and is a single digit from 2 to 4 otherwise; we also verified this for every radicand from 1 to 999,999. Division scales by $f = \lfloor 10/(d_0+1) \rfloor \in \{1, 2, 3, 5\}$, and $f (d_0 + 1) \le 10$ keeps the scaled divisor at the same length. The scalings use the multiplication procedure; the remainder of a division is unscaled by multiplying by $10/f$ and dropping the last digit ($f = 2, 5$) or by an exact division by 3, and the root by multiplying by $10/c$ or $100/c$ and dropping one or two digits. The runtime's choice of scale factors and initial guesses affects the number of steps but not the result, since every guess is at least the true digit and is lowered only by certified steps. The cost model was fitted on batches of 1 to 32 views on the M4.

The browser test drew its 200 problems from a fixed generator before any run: $a \div b$ to 10 places with $a \le 999$ and $2 \le b \le 99$; $\sqrt a$ to 6 places with $2 \le a \le 999$; $\gcd(a, b)$ with $10 \le a, b \le 999$; and signed sums and differences of 1 to 20 digits. It ran in Chrome on an Apple M4, one problem at a time, with independent views of the same problem sharing a forward pass; every view was a table row. Answers were checked against arbitrary-precision arithmetic. The test was repeated with the gated adapter, with identical views and outputs.

\subsection*{Algebra}
The algebra runtime parses a problem, constructs the derivation and obtains every number through the calculator interface; independent calls are batched. Each line of the derivation records the model calls that produced its numbers. The default check substitutes rational roots into the collected equation and evaluates expansions at $x = 1$; the strict check substitutes into the original expressions and evaluates expansions at $x = -1, 0, 1, 2$. For equations, both modes also confirm the collection of terms by evaluating each side before and after collection at $x = y = z = 1$; this does not detect a coefficient moved between variables, which the strict mode's substitution into the original equations does. Evaluation at fixed points is a spot check: an incorrect expansion of degree $D$ passes if the difference from the correct one vanishes at every checked point, which a single wrong coefficient cannot achieve at $x = 1$. For an irrational pair, the runtime removes square factors from the discriminant by trial division by the primes up to 47 and divides $p$, $q$ and $d$ by their greatest common divisor, so the radical is fully reduced when the discriminant is below $53^2 = 2{,}809$. The pair, written $(p \pm q\sqrt m)/d$ with an integer $p$ and positive integers $q$, $m$ and $d$, is accepted only if
\renewcommand{\theHequation}{eq-vieta}
\begin{equation}
2ap = -bd \quad \text{and} \quad a\,(p^2 - q^2 m) = c\,d^2,
\tag{11}\label{eq-vieta}
\end{equation}
which are Vieta's relations $x_1 + x_2 = -b/a$ and $x_1 x_2 = c/a$ multiplied by $ad$ and $ad^2$; for $a \neq 0$ they hold if and only if the pair are the two roots. The check concerns the values of the roots; that the form is reduced rests on the reduction above. Fault injection planted an incorrect root (200 problems per route) or corrupted one arithmetic result (200 per route), with fixed seeds, and the rejection rates are the same in both check modes; a further 200 perturbed radical pairs were all rejected. The reference test suite (264 tests, 2,200 generated problems per route and check mode) uses exact tables in place of the model.

\subsection*{Gating}
The gate multiplies each adapter branch before it is added to the frozen projection. At $g = 0$ the branches are still evaluated, which adds the adapter's 1.2\% of parameters to each forward pass. The identity test compared the base, ungated and gated graphs bit for bit under ONNX Runtime 1.31.1 (the Python package) on CPU, on 100 chat prompts (first user turns of the No Robots test split\cite{norobots2023}; one prefill and 24 greedy steps each, 2,500 forward passes) and on 2,688 table views: the 2,446 views and the 242 addition views again in the padded prompt form. Every logit, hidden state and cached state of the gated graph equals that of the base model at $g = 0$ and that of the ungated adapter at $g = 1$, and greedy completions of the 100 chat prompts are identical.

\subsection*{Dialogue fine-tuning}
The training data comprise about 3,150 synthetic dialogues, about 30\% of which should not invoke the sheet, and 547 problems from the GSM8K training split, solved by the base model with each calculation replaced by a call to the sheet. A call is a Markdown code block labelled \texttt{sheet}, with one calculator expression or equation per line; generation stops when the block closes, the runtime executes the block and appends one result line per expression, and generation resumes, with at most two blocks per reply. Every training step, in the supervised, distillation and reinforcement-learning stages alike, includes a batch of table rows under the objective (\hyperref[eq-loss]{9}), with low-margin rows replayed, and every stage was required to end with all 2,446 rows passing. The held-out set contains 780 items from unseen templates. Invocations are scored by execution. Criteria were fixed before training: precision and recall of invocation of at least 0.95; correct invocations for at least 90\% of word problems and 95\% of equations; IFEval, GSM8K (250 test problems) and MMLU (500 test questions; both subsets drawn with seed 0) within 2 points of the base model. GSM8K is evaluated at $g = 1$ with the sheet in the loop; in these evaluations, and in the reward, the blocks are executed by a reference implementation of the procedures in exact integer arithmetic rather than by the model, which by Theorem \hyperref[thm-p]{1} returns the same results. The dialogue adapter was the fifth candidate scored on the same benchmark items. We therefore rescored it and the base model on the 1,069 GSM8K test problems and 1,000 MMLU test questions not used before, with predictions and a stopping rule fixed in advance (GSM8K \ensuremath{-}1.5 within [\ensuremath{-}4.5, +1.5]; MMLU \ensuremath{-}1.0 within [\ensuremath{-}3.0, +1.0]; failure below \ensuremath{-}2 by more than one standard error). GSM8K fell outside its band: 0.491 against 0.549 (139 problems gained, 201 lost). The adapter called the sheet on 619 problems and solved 327, where the base model solved 368 of the same; on the other 450 it solved 198 against 219. MMLU stayed inside its band, at 0.463 against 0.483. The benchmark differences are reported with the standard error of a paired difference, $\sqrt{n_{+} + n_{-}}/n$ for $n_{+}$ and $n_{-}$ items gained and lost out of $n$. As an alternative to training the decision, we switched the gate within a reply: the base model, prompted with four examples, writes the prose at $g = 0$, and an adapter writes only the block at $g = 1$; two preliminary dialogue adapters served as block writers (Fig. \hyperref[fig-gate]{5}b,c). The switch is exact, but invocation inherits the base model's recall of 0.809, and the sheet prompt lowered the base model's GSM8K accuracy by 34 to 37 points. Reinforcement learning used GRPO with groups of eight samples per task and four tasks per step, a reward that requires a finished reply quoting the executed result and penalises repeated or excess invocations, and a KL penalty toward the supervised checkpoint on every generated token; it ran for 200 steps on one H100 GPU.

\subsection*{Mechanism}
The analyses use the arithmetic adapter. Depth $\ell$ counts layers from the input: $h_0$ is the embedding and $h_\ell$ the output of the $\ell$-th of the 24 layers, so the full-attention layers end at depths 4, 8, \ldots{}, 24. The logit lens (\hyperref[eq-lens]{7}) is computed at every depth at the first output position, for the 1,795 addition and multiplication views. The decision depth of a view is the first depth after which the final output remains the top token. Fig. \hyperref[fig-inside]{6}b projects the final-layer states, after the final normalisation and with the table's mean subtracted, onto an orthonormal basis of the three leading Fisher discriminant directions of the output classes, with the within-class covariance shrunk 90\% toward a multiple of the identity; the nearest-centre shares are measured in this projection. The class-mean intervention (\hyperref[eq-steer]{8}) acts at the first output position on 80 views per table drawn without replacement with a fixed seed, excluding the halting view of \texttt{SUM} and the \texttt{MUL} views with a blank multiplicand digit, and the target is drawn uniformly from the table's other outputs; the class means are taken over the same eligible views (241 \texttt{SUM} views and the 1,000 \texttt{MUL} views with a multiplicand digit). The projection of Fig. \hyperref[fig-inside]{6}b uses all 242 \texttt{SUM} and 1,100 \texttt{MUL} views.

\subsection*{Baselines}
The comparison was pre-registered with its scoring rule. Problems were generated with a fixed seed; every model received the prompt ``What is a + b? Give the exact integer. Put the final answer in \textbackslash{}boxed\{\}.'' (with \ensuremath{\times} for products). The answer was the last boxed integer, and a reply that exhausted the budget or gave no answer was scored as wrong. Qwen3.5-0.8B and Qwen3.5-27B (FP8 release) ran under vLLM with the recommended sampling settings and budgets of 32,768 tokens with chain of thought and 4,096 without. GPT-6.1 Sol (\texttt{gpt-6.\allowbreak{}1-sol}, reasoning effort high), Claude Opus 5.5 (\texttt{claude-opus-5-5}, effort high) and Gemini 4 Argon (\texttt{gemini-4-argon}, thinking level high) were accessed through their APIs on 31 October 2026 with default sampling and a budget of 32,768 output tokens including reasoning. Five open-weight models, DeepSeek-V4-Pro (\texttt{deepseek-v4-pro:0813}), Mistral-Large-4, Gemma 4 31B, GLM-5.3 and gpt-oss-120b, ran through the Ollama cloud API on 31 October 2026 with reasoning enabled at the highest available setting and the same budget; Kimi K3 was piloted and excluded on cost. Each model was run once per problem, so a difference of a few problems between two models is within sampling variation. Sketchpad ran all 220 problems in batched lockstep in PyTorch, 64,700 views in 1,001 forward passes.

\subsection*{Related work}\label{sec-related}
\textbf{Neural arithmetic.} The Neural GPU learned binary addition and multiplication that generalise beyond the training length\cite{kaiser2016neuralgpu,price2016neuralgpu}, and specialised arithmetic units followed\cite{trask2018nalu,madsen2020nau}. For transformers, positional encodings and data formats determine length generalisation\cite{nogueira2021limitations,anil2022length,lee2024teaching,zhou2024raspl,zhou2024robust,mcleish2024abacus,cho2024coupling}, and fine-tuned models reach high accuracy on bounded arithmetic\cite{liu2023goat,yang2023mathglm}, with division and gcd remaining difficult\cite{shen2024revorder,charton2024gcd,kasliwal2026division}. Neural cellular automata and differentiable transducers trained from scratch add and multiply far beyond their training lengths\cite{wei2026nca,wei2026ncaadd,dfst2025}. Mechanistic analyses find addition circuits correct on almost all inputs\cite{quirke2024verified}. Exactness has been proved under idealised training\cite{backdeluca2025exact}, for small recurrent networks that add inputs of any length after training with a description-length objective\cite{lan2022mdl}, and for a purpose-built transformer parameterisation that evaluates Boolean circuits of any depth\cite{ito2026circuits}. Sketchpad differs in using a pretrained vision-language model, in covering division, square roots and gcd, and in certifying the trained weights by enumeration, where these works estimate accuracy on samples or prove exactness for idealised, small or purpose-built networks.

\textbf{Procedures in language models.} Neural programmer-interpreters learned grade-school addition as a sequence of steps on a scratch pad read through pointers\cite{reed2016npi}, the architecture whose generalisation Cai et al. later verified\cite{cai2017recursion}, and a controller moving over a two-dimensional grid learned multi-digit addition that extends to thousands of digits\cite{zaremba2016simple}. CAEF fine-tunes Llama-3.1-8B to execute arithmetic operators step by step on a text tape, with one adapter per operator\cite{lai2024caef}. Other work embeds non-learned arithmetic modules\cite{dugan2024occamllm,igc2025}, compiles programs into transformer weights\cite{giannou2023looped,loom2026,percepta2026vm,dong2023neuralcomprehension}, trains transformers to predict program execution traces\cite{xu2026universal}, or lets a language model emulate an interpreter\cite{li2024chainofcode}.

\textbf{Tool use.} Toolformer\cite{schick2023toolformer}, PAL, Program of Thoughts, ToRA, ReTool and ToolRL\cite{gao2023pal,chen2023pot,gou2024tora,feng2025retool,qian2025toolrl} and recursive language models\cite{zhang2025rlm} delegate computation to external interpreters.

\textbf{Verification.} Neural network verification typically bounds outputs over continuous input regions\cite{katz2017reluplex}; here the input set is finite, as in computer-assisted proofs of a small transformer's accuracy on a task of fixed input size\cite{gross2024compact}, except that a finite set of views covers inputs of every length. Our theorems specialise results on the identification of finite machines\cite{moore1956gedanken,chow1978testing}, the verification of learned recursive programs\cite{cai2017recursion} and certified universal computation by language models\cite{schuurmans2024universal,lewandowski2026universal}. Lewandowski et al. likewise restrict each decoding step to a fixed prompt and a local pair of symbols, and check a pretrained model's output on all 1,857 rules of a universal system by exact match at temperature zero; Sketchpad trains a small model to meet the condition and certifies margins on the deployment path, which Theorem \hyperref[thm-transfer]{3} makes robust to that path's numerical differences. Discrete neural algorithmic reasoning forces a processor through finite predefined states, so that its learned algorithms can be proved correct on any input\cite{rodionov2025dnar}; it trains processors from scratch, where Sketchpad certifies a pretrained model. The context discipline resembles the inductive scratchpad\cite{abbe2024globality} and PENCIL\cite{yang2025pencil}. Certification from examples can be hard in general and length-generalisation bounds can be uncomputable\cite{backdeluca2026certification,yang2026lengthbounds}; the pinned-sheet policy avoids these obstacles by making the input set finite.

\textbf{Visual input and scratchpads.} Pixel-based language models\cite{rust2023pixel,tschannen2023clippo,lee2023pix2struct} and visual context compression\cite{wei2025deepseekocr,cheng2025glyph} read rendered text. A small network learned addition and subtraction of seven-digit numbers, but not multiplication, from pictures of numbers\cite{hoshen2016visual}, and vision-language models still fail at multiplication from images\cite{balter2026vlm}. Visual scratchpads support multimodal reasoning\cite{hu2024visualsketchpad,menon2024whiteboard,li2025mvot,selfimagine2024,mathcanvas2025}. Visual Sketchpad\cite{hu2024visualsketchpad} shares the name of our system, which refers to Sutherland's Sketchpad\cite{sutherland1963sketchpad}.

\textbf{Algebra.} Transformers have been trained for symbolic integration\cite{lample2020symbolic}, polynomial simplification\cite{agarwal2021polynomial} and Gr\"{o}bner bases\cite{kera2024groebner}, and large models solve school algebra with chain of thought\cite{lewkowycz2022minerva,feng2023cot}. None of these is exact by construction.

\textbf{Continual learning.} Learning without Forgetting\cite{li2017lwf}, progressive networks\cite{rusu2016progressive}, elastic weight consolidation\cite{kirkpatrick2017ewc} and adapter routing\cite{buehler2024xlora} address forgetting; on-policy distillation and the analysis of Shenfeld et al. relate forgetting to divergence from the base model on its own outputs\cite{agarwal2024gkd,lu2025onpolicy,shenfeld2025razor}. Our gate is a hard switch in the spirit of progressive networks. Activated LoRA likewise applies an adapter only to tokens after its invocation, reusing the base model's cache for earlier tokens\cite{greenewald2025alora}, and modality-gated adapters leave inputs of other modalities bit-for-bit unchanged, as their authors verify by exact-equality tests\cite{tonmoy2026modalitygated}. In Sketchpad the gate is a graph input set by the runtime.

\subsection*{Data and code availability}
The model, adapter, certificate and analysis data are available on Hugging Face (\texttt{mohsenvand/\allowbreak{}sketchpad-0.\allowbreak{}8b-preview}). The runtime, the algebra, the baselines and the sources of this paper are available under the Apache 2.0 licence at https://github.com/NeoVand/sketchpad. The Lean proofs are in \texttt{paper/\allowbreak{}lean/\allowbreak{}} of the same repository (Appendix \hyperref[app-proofs]{A}).

\section*{Extended Data}\label{extended}
\begin{table}[tbp]
\centering
\caption*{\textbf{Extended Data Table 1 $|$} \textbf{Tables.} Each row is one view an operation can encounter. Margins are the minimum over the table, in logits, on the server (PyTorch, 32-bit) and on the M4 (WebGPU, 16-bit), for the arithmetic adapter. \texttt{SUM} has one output position ($K_v = 1$), the other operations three; \textendash{} marks an output that is always the same placeholder. The five multiplication tables, \texttt{MUL} to \texttt{NXT}, total 1,553 views. Next-operation codes: \texttt{A} add the partial product, \texttt{Z} end of pass, \texttt{M} multiply (or, after \texttt{EST}, proceed with the guess), \texttt{F} flush, \texttt{N} next digit (after \texttt{EST}, the guess is 0), \texttt{H} halt, \texttt{B} final borrow, \texttt{S} continue subtraction, \texttt{C} correct the remainder.}\phantomsection\label{tbl-tables}
{\small
\begin{tabularx}{\linewidth}{l>{\raggedright\arraybackslash}X>{\raggedright\arraybackslash}Xrrr}
\toprule
Operation & View (top to bottom) & Outputs at \texttt{=\allowbreak{}}, \texttt{?}, \texttt{?} & Rows & Server & M4 \\
\midrule
\texttt{SUM} & carry (blank, 1), top digit (blank, 0\textendash{}9), bottom digit (blank, 0\textendash{}9) & sum and carry as one token (0\textendash{}19), or \texttt{H} & 242 & 3.96 & 3.02 \\
\texttt{MUL} & multiplicand digit (blank, 0\textendash{}9), multiplier digit (0\textendash{}9), carry (blank, 0\textendash{}8) & digit, carry, \texttt{A} or \texttt{Z} & 1,100 & 6.39 & 6.52 \\
\texttt{ADD} & carry (blank, 1), product digit (0\textendash{}9), running-total digit (blank, 0\textendash{}9) & digit, carry, \texttt{M} & 220 & 9.37 & 9.50 \\
\texttt{ADZ} & as \texttt{ADD}, at the end of a pass & digit, carry, \texttt{F} & 220 & 8.41 & 8.00 \\
\texttt{FLS} & carry (blank, 1), blank, blank & final carry, \textendash{}, \texttt{N} & 2 & 12.93 & 13.11 \\
\texttt{NXT} & next multiplier digit (blank, 0\textendash{}9), blank, blank & \textendash{}, \textendash{}, \texttt{M} or \texttt{H} & 11 & 10.48 & 10.22 \\
\texttt{SUB} & borrow (blank, 1), top digit (blank, 0\textendash{}9), bottom digit (blank, 0\textendash{}9) & digit, borrow, \texttt{S}; top and bottom blank: \textendash{}, \textendash{}, \texttt{H} (no borrow) or \texttt{B} (borrow) & 242 & 8.36 & 8.19 \\
\texttt{EST} & leading remainder digits $r_1$ (0\textendash{}$d$) and $r_2$ (0\textendash{}9), leading divisor digit $d$ (5\textendash{}9) & $\hat q = \min(9, \lfloor(10r_1+r_2)/d\rfloor)$, \textendash{}, \texttt{M} if $\hat q > 0$, else \texttt{N} & 400 & 4.51 & 4.61 \\
\texttt{DEC} & estimate $q$ (1\textendash{}9), blank, blank & $q-1$, \textendash{}, \texttt{C} & 9 & 10.00 & 10.09 \\
\textbf{All} &  &  & \textbf{2,446} & \textbf{3.96} & \textbf{3.02} \\
\bottomrule
\end{tabularx}
}
\end{table}
\begin{table}[tbp]
\centering
\caption*{\textbf{Extended Data Table 2 $|$} \textbf{Browser test.} 50 problems per set on the Apple M4 (WebGPU, 16-bit decoder), one problem at a time. Passes are fewer than views because independent views of one problem, such as the \texttt{MUL} and \texttt{SUB} views of a multiply-subtract pass, share a forward pass.}\phantomsection\label{tbl-t3}
{\small
\begin{tabularx}{\linewidth}{>{\raggedright\arraybackslash}Xrrrrrr}
\toprule
Set & Exact & Median s & Max s & Over 12 s & Median passes / views & Cost model median s \\
\midrule
$a \div b$ to 10 places ($a \le 999$, $b \le 99$) & 50 & 7.0 & 13.3 & 1 & 65 / 90 & 7.6 \\
$\sqrt a$ to 6 places ($a \le 999$) & 50 & 9.8 & 16.5 & 11 & 86 / 137 & 10.7 \\
$\gcd(a,b)$ ($a, b \le 999$) & 50 & 7.0 & 13.4 & 4 & 61 / 84 & 7.1 \\
signed $\pm$, up to 20 digits & 50 & 2.0 & 4.7 & 0 & 19 / 19 & 1.9 \\
\bottomrule
\end{tabularx}
}
\end{table}
\begin{table}[tbp]
\centering
\caption*{\textbf{Extended Data Table 3 $|$} \textbf{Cost of the algebra benchmark.} Views, forward passes after batching and seconds under the cost model, default check, for the textbook and evaluation routes. The first eleven problems are those of Fig. \hyperref[fig-algebra]{4}c. The last three have irrational roots, which the textbook route returns as radicals and the evaluation route only as decimal bounds, so they have no comparable evaluation entry and were not measured on the M4.}\phantomsection\label{tbl-algebra}
{\small
\begin{tabularx}{\linewidth}{>{\raggedright\arraybackslash}Xrr}
\toprule
Problem & Textbook views / passes / s & Evaluation views / passes / s \\
\midrule
\texttt{solve 3(x-2)+5=\allowbreak{}2x+14} & 36 / 22 / 2.8 & 75 / 27 / 4.5 \\
\texttt{solve 7x+13=\allowbreak{}2026-4x} & 97 / 62 / 7.6 & 165 / 88 / 11.8 \\
\texttt{solve 2x+3y=\allowbreak{}7,\allowbreak{} x-y=\allowbreak{}1} & 75 / 48 / 5.9 & 194 / 41 / 9.8 \\
\texttt{solve (2x-3)(x+5)=\allowbreak{}x\textasciicircum{}2+1-16} & 79 / 43 / 5.7 & 177 / 50 / 9.8 \\
\texttt{solve x\textasciicircum{}2-x-6=\allowbreak{}0} & 102 / 77 / 8.8 & 162 / 79 / 11.1 \\
\texttt{123456 * 654321} & 96 / 46 / 6.5 & 96 / 46 / 6.5 \\
\texttt{expand (x+2)\textasciicircum{}5} & 159 / 95 / 12.1 & 458 / 248 / 33.1 \\
\texttt{expand (2x-3)\textasciicircum{}4} & 172 / 94 / 12.5 & 492 / 274 / 36.0 \\
\texttt{factor x\textasciicircum{}3-6x\textasciicircum{}2+11x-6} & 195 / 149 / 16.9 & 483 / 210 / 31.5 \\
\texttt{expand (x+1)\textasciicircum{}10} & 322 / 154 / 21.9 & 2,359 / 845 / 142.2 \\
\texttt{solve 2x+3y-z=\allowbreak{}1,\allowbreak{} x-y+2z=\allowbreak{}3,\allowbreak{} 3x+y+z=\allowbreak{}8} & 323 / 167 / 22.8 & 538 / 102 / 26.5 \\
\texttt{solve x\textasciicircum{}2-2=\allowbreak{}0} & 86 / 65 / 7.4 &  \\
\texttt{solve 2x\textasciicircum{}2+3x-1=\allowbreak{}0} & 144 / 88 / 11.0 &  \\
\texttt{solve x\textasciicircum{}2-6x+4=\allowbreak{}0} & 133 / 96 / 11.2 &  \\
\bottomrule
\end{tabularx}
}
\end{table}
\begin{table}[tbp]
\centering
\caption*{\textbf{Extended Data Table 4 $|$} \textbf{Dialogue fine-tuning.} Held-out templates with the gate on and no examples, against the base model prompted with four examples (invocation and translation) or with none (benchmarks). IFEval, GSM8K and MMLU in points relative to the base model (0.536, 0.516 and 0.476), on the same 541, 250 and 500 items for every recipe; for the dialogue adapter, with the standard error of the paired difference. The last row rescores the dialogue adapter on GSM8K and MMLU items not used before (Methods). GSM8K replies at $g = 1$ invoke the sheet, executed by the reference implementation (Methods); \textendash{}, not measured.}\phantomsection\label{tbl-chat}
{\small
\begin{tabularx}{\linewidth}{>{\raggedright\arraybackslash}Xlrrrrrr}
\toprule
Recipe & Precision / recall & Word & Equations & Prose & IFEval & GSM8K & MMLU \\
\midrule
criterion & \ensuremath{\ge} 0.95 / \ensuremath{\ge} 0.95 & \ensuremath{\ge} 0.90 & \ensuremath{\ge} 0.95 & \ensuremath{\ge} 0.95 & \ensuremath{\ge} \ensuremath{-}2 & \ensuremath{\ge} \ensuremath{-}2 & \ensuremath{\ge} \ensuremath{-}2 \\
base model, four examples & 0.971 / 0.809 & 0.789 & 0.583 & 0.667 & 0 & 0 & 0 \\
synthetic dialogues, self-distillation (from a preliminary multiplication adapter) & 0.957 / 0.996 & 0.872 & 0.967 & 0.983 & \ensuremath{-}4.8 & \textendash{} & \ensuremath{-}6.6 \\
synthetic dialogues, self-distillation & 1.000 / 0.998 & 0.894 & 0.975 & 1.000 & \ensuremath{-}0.55 & \ensuremath{-}31.2 & \ensuremath{-}5.4 \\
+ human multi-step solutions & 1.000 / 0.994 & 0.856 & 0.975 & 0.992 & \ensuremath{-}2.8 & \ensuremath{-}14.0 & \ensuremath{-}2.4 \\
+ base-model multi-step solutions & 1.000 / 0.994 & 0.817 & 0.958 & 0.992 & \ensuremath{-}4.4 & \ensuremath{-}6.4 & \ensuremath{-}3.4 \\
+ on-policy distillation & 1.000 / 0.946 & 0.678 & 0.958 & 0.983 & \ensuremath{-}1.5 & \ensuremath{-}2.0 & \ensuremath{-}2.2 \\
+ wider word-problem data & 1.000 / 0.991 & 0.856 & 0.958 & 0.975 & +0.2 & \ensuremath{-}2.4 & \ensuremath{-}2.2 \\
+ GRPO & 1.000 / 0.981 & 0.861 & 0.958 & 0.992 & \ensuremath{-}2.0 & \ensuremath{-}4.8 & \ensuremath{-}0.6 \\
+ GRPO, strict reward and task mix & 1.000 / 0.794 & 0.378 & 0.933 & 0.975 & \ensuremath{-}2.2 & \ensuremath{-}0.8 & \ensuremath{-}1.0 \\
\textbf{+ GRPO anchored to the supervised checkpoint (dialogue adapter)} & \textbf{0.994 / 0.989} & \textbf{0.928} & \textbf{0.967} & \textbf{0.983} & \textbf{\ensuremath{-}1.5 \ensuremath{\pm} 2.0} & \textbf{\ensuremath{-}1.6 \ensuremath{\pm} 3.4} & \textbf{\ensuremath{-}1.4 \ensuremath{\pm} 1.6} \\
dialogue adapter, fresh items (1,069 GSM8K, 1,000 MMLU) & \textendash{} & \textendash{} & \textendash{} & \textendash{} & \textendash{} & \textbf{\ensuremath{-}5.8 \ensuremath{\pm} 1.7} & \textbf{\ensuremath{-}2.0 \ensuremath{\pm} 1.1} \\
\bottomrule
\end{tabularx}
}
\end{table}
\begin{table}[tbp]
\centering
\caption*{\textbf{Extended Data Table 5 $|$} \textbf{Comparison.} Problems answered exactly, of 20 per cell; in brackets, the number of replies that exhausted the token budget. Qwen3.5-27B is the FP8 release; all other models ran with reasoning at their highest available setting, accessed on 31 October 2026. The errors of GPT-6.1 Sol and Claude Opus 5.5 were completed replies well within the budget, differing from the correct sum in 1 to 637 digits; 9 of the 13 errors of Gemini 4 Argon exhausted the budget. Most errors of the open-weight models exhausted the budget, except those of Gemma 4 31B, 46 of whose 64 errors were incorrect answers in completed replies. Qwen3.5-27B exhausted the budget in 93 of its 99 failures, and Qwen3.5-0.8B with chain of thought in all of its 217.}\phantomsection\label{tbl-baselines}
{\small
\footnotesize\setlength{\tabcolsep}{3pt}
\begin{tabularx}{\linewidth}{>{\raggedright\arraybackslash}Xrrrrrrrrrrrr}
\toprule
System & +10 & +30 & +100 & +300 & +1,000 & \ensuremath{\times}2 & \ensuremath{\times}4 & \ensuremath{\times}6 & \ensuremath{\times}10 & \ensuremath{\times}15 & \ensuremath{\times}20 & All \\
\midrule
Sketchpad & 20 & 20 & 20 & 20 & 20 & 20 & 20 & 20 & 20 & 20 & 20 & 220 \\
GPT-6.1 Sol & 20 (0) & 20 (0) & 20 (0) & 18 (0) & 19 (0) & 20 (0) & 20 (0) & 20 (0) & 20 (0) & 20 (0) & 20 (0) & 217 \\
Claude Opus 5.5 & 20 (0) & 20 (0) & 20 (0) & 19 (0) & 17 (0) & 20 (0) & 20 (0) & 20 (0) & 20 (0) & 20 (0) & 20 (0) & 216 \\
Gemini 4 Argon & 20 (0) & 20 (0) & 20 (0) & 18 (1) & 9 (8) & 20 (0) & 20 (0) & 20 (0) & 20 (0) & 20 (0) & 20 (0) & 207 \\
DeepSeek-V4-Pro & 20 (0) & 20 (0) & 19 (0) & 18 (0) & 2 (17) & 20 (0) & 20 (0) & 20 (0) & 20 (0) & 20 (0) & 20 (0) & 199 \\
Mistral-Large-4 & 20 (0) & 20 (0) & 14 (5) & 7 (10) & 0 (20) & 20 (0) & 20 (0) & 20 (0) & 20 (0) & 20 (0) & 7 (13) & 168 \\
Gemma 4 31B & 20 (0) & 20 (0) & 17 (0) & 6 (2) & 0 (15) & 20 (0) & 20 (0) & 20 (0) & 19 (1) & 9 (0) & 5 (0) & 156 \\
GLM-5.3 & 20 (0) & 19 (1) & 11 (6) & 4 (15) & 0 (20) & 20 (0) & 20 (0) & 20 (0) & 20 (0) & 16 (3) & 2 (18) & 152 \\
gpt-oss-120b & 20 (0) & 20 (0) & 10 (9) & 0 (18) & 0 (20) & 20 (0) & 20 (0) & 20 (0) & 20 (0) & 16 (4) & 3 (17) & 149 \\
Qwen3.5-27B, chain of thought & 18 (2) & 12 (8) & 8 (10) & 1 (19) & 0 (20) & 20 (0) & 20 (0) & 15 (4) & 14 (6) & 6 (11) & 7 (13) & 121 \\
Qwen3.5-0.8B, chain of thought & 0 (20) & 0 (20) & 0 (20) & 0 (20) & 0 (20) & 3 (17) & 0 (20) & 0 (20) & 0 (20) & 0 (20) & 0 (20) & 3 \\
Qwen3.5-0.8B & 0 (8) & 0 (20) & 0 (20) & 0 (20) & 0 (20) & 15 (4) & 8 (9) & 0 (14) & 0 (20) & 0 (19) & 0 (20) & 23 \\
\bottomrule
\end{tabularx}
}
\end{table}
\appendix
\section{Proofs}\label{app-proofs}
The theorems are proved in the Lean 4 proof assistant\cite{demoura2021lean4} with Mathlib\cite{mathlib2020} (version 4.35.0), without \texttt{sorry} or additional axioms; \texttt{\#print axioms} reports only \texttt{propext}, \texttt{Classical.\allowbreak{}choice} and \texttt{Quot.\allowbreak{}sound}, or a subset. The formalisation models the network as a deterministic function of its context, or as a causal decoder whose layer at a position reads only earlier positions, with arbitrary tie-breaking in the argmax and exact arithmetic; floating point enters only through $\tau$ in Theorem \hyperref[thm-transfer]{3}. The proofs below follow the Lean proofs, whose declarations are listed in Extended Data Table \hyperref[tbl-lean]{6} (in the namespace \texttt{Sketchpad.\allowbreak{}Pinned} unless prefixed). Lemmas \hyperref[lem-argmax]{1}\textendash{}\hyperref[lem-cache]{4} and Theorem \hyperref[thm-compose]{4}a are formalised in general, and Theorems \hyperref[thm-p]{1}, \hyperref[thm-compose]{4}b and \hyperref[thm-transfer]{3} for the sheet machine (\texttt{Sheet.\allowbreak{}lean}) and for a token form, in which the state is a token and each step reads a state token and an event token after a fixed prompt (\texttt{Pinned.\allowbreak{}lean}, \texttt{Compose.\allowbreak{}lean}). In the sheet form of Theorem \hyperref[thm-compose]{4}b, each call runs a sheet procedure for a given number of steps from a configuration built from its input and reads its result from the final configuration. The hypotheses $R_T(I) \subseteq V$ and the correctness of the reference procedures are assumptions of these theorems, not formalised for the concrete layouts. \texttt{Paper.\allowbreak{}lean} names every statement by its number in this paper.

\begin{table}[tbp]
\centering
\caption*{\textbf{Extended Data Table 6 $|$} \textbf{Lean statements.} The package is in \texttt{paper/\allowbreak{}lean/\allowbreak{}} of the Sketchpad repository (Lean 4.35.0, Mathlib); the files other than \texttt{Sheet.\allowbreak{}lean} and \texttt{Paper.\allowbreak{}lean} are in its \texttt{Sketchpad/\allowbreak{}} folder. \texttt{Paper.\allowbreak{}lean} provides aliases by paper number, such as \texttt{theorem2\_\allowbreak{}counterexample} for \texttt{theorem8a\_\allowbreak{}counterexample}.}\phantomsection\label{tbl-lean}
{\small
\begin{tabularx}{\linewidth}{l>{\raggedright\arraybackslash}X>{\raggedright\arraybackslash}X}
\toprule
Statement & Lean declarations & File \\
\midrule
Theorem \hyperref[thm-p]{1} & \texttt{Sketchpad.\allowbreak{}sheet\_\allowbreak{}exact\_\allowbreak{}iff}, \texttt{Sketchpad.\allowbreak{}sheet\_\allowbreak{}certified}; token form \texttt{regeneration\_\allowbreak{}trace\_\allowbreak{}iff}, \texttt{pinned\_\allowbreak{}regeneration\_\allowbreak{}iff} & \texttt{Sheet.\allowbreak{}lean}, \texttt{Pinned.\allowbreak{}lean} \\
Theorem \hyperref[thm-nofinite]{2} & \texttt{no\_\allowbreak{}finite\_\allowbreak{}test\_\allowbreak{}suite}, \texttt{theorem8a\_\allowbreak{}counterexample}, \texttt{no\_\allowbreak{}finite\_\allowbreak{}check} & \texttt{Unpinned.\allowbreak{}lean} \\
Theorem \hyperref[thm-transfer]{3} & \texttt{Sketchpad.\allowbreak{}sheet\_\allowbreak{}exact\_\allowbreak{}of\_\allowbreak{}close}; token form \texttt{regeneration\_\allowbreak{}of\_\allowbreak{}close}; unpinned decoding \texttt{unpinned\_\allowbreak{}exact\_\allowbreak{}of\_\allowbreak{}close} & \texttt{Sheet.\allowbreak{}lean}, \texttt{Pinned.\allowbreak{}lean}, \texttt{Unpinned.\allowbreak{}lean} \\
Theorem \hyperref[thm-compose]{4} & (a) \texttt{error\_\allowbreak{}decomposition}; (b) \texttt{Sketchpad.\allowbreak{}sheet\_\allowbreak{}compose\_\allowbreak{}exact}, \texttt{Sketchpad.\allowbreak{}sheet\_\allowbreak{}error\_\allowbreak{}bound}; token form \texttt{compose\_\allowbreak{}exact}, \texttt{error\_\allowbreak{}bound} & \texttt{Compose.\allowbreak{}lean}, \texttt{Sheet.\allowbreak{}lean} \\
Lemma \hyperref[lem-argmax]{1} & \texttt{sel\_\allowbreak{}eq\_\allowbreak{}of\_\allowbreak{}close}, \texttt{sel\_\allowbreak{}eq\_\allowbreak{}of\_\allowbreak{}strict} & \texttt{Pinned.\allowbreak{}lean} \\
Lemma \hyperref[lem-regen]{2} & \texttt{Sketchpad.\allowbreak{}Interp.\allowbreak{}regeneration}, \texttt{scanl\_\allowbreak{}eq\_\allowbreak{}iff\_\allowbreak{}foldl\_\allowbreak{}eq}, \texttt{regeneration\_\allowbreak{}trace\_\allowbreak{}iff} & \texttt{Interp.\allowbreak{}lean}, \texttt{Pinned.\allowbreak{}lean} \\
Lemma \hyperref[lem-hist]{3} & \texttt{histRun\_\allowbreak{}eq\_\allowbreak{}of\_\allowbreak{}closed}, \texttt{reach\_\allowbreak{}closed} & \texttt{Pinned.\allowbreak{}lean} \\
Lemma \hyperref[lem-cache]{4} & \texttt{resid\_\allowbreak{}congr}, \texttt{cache\_\allowbreak{}sound}, \texttt{cachedForward\_\allowbreak{}eq}, \texttt{policy\_\allowbreak{}tick\_\allowbreak{}eq} & \texttt{Pinned.\allowbreak{}lean}, \texttt{Policy.\allowbreak{}lean} \\
\bottomrule
\end{tabularx}
}
\end{table}
\subsection*{Definitions and lemmas}
\begin{thmbox}\phantomsection\label{def-machine}\noindent\textbf{Definition 1} (Sheet machine).\enspace\itshape A sheet machine consists of sets $C$ of configurations, $\mathcal V$ of views and $A$ of outputs, a view map $\mathrm{view}\colon C \to \mathcal V \cup \{\bot\}$ and a paint map $\mathrm{paint}\colon C \times A \to C$. For $g\colon \mathcal V \to A$, the step $\sigma_g\colon C \to C$ is
\renewcommand{\theHequation}{eq-sigma}
\begin{equation}
\sigma_g(c) = \begin{cases} \mathrm{paint}\big(c, g(v)\big) & \text{if } \mathrm{view}(c) = v \in \mathcal V, \\ c & \text{if } \mathrm{view}(c) = \bot, \end{cases}
\tag{12}\label{eq-sigma}
\end{equation}
and the run of length $n$ from $c_0$ is $(c_0, \sigma_g(c_0), \dots, \sigma_g^n(c_0))$. The model's run and the reference run are those of $\sigma_f$ and $\sigma_T$. For $I \subseteq C$,
\[
\mathrm{Reach}_T(I) = \{\sigma_T^n(c_0) : c_0 \in I,\ n \ge 0\}, \qquad R_T(I) = \{\mathrm{view}(c) : c \in \mathrm{Reach}_T(I)\} \setminus \{\bot\}.
\]\upshape
\end{thmbox}
\begin{thmbox}\phantomsection\label{def-fold}\noindent\textbf{Definition 2} (Runs over events).\enspace\itshape Let $S$ be a set of states, $E$ a set of events and $F\colon S \times E \to S$. For $s_0 \in S$ and a word $w$ over $E$, $F^w(s_0)$ is the left fold, $F^{\varepsilon}(s_0) = s_0$ and $F^{ue}(s_0) = F(F^u(s_0), e)$, and the trace of $F$ on $w = e_1 \cdots e_n$ is $\big(F^{e_1 \cdots e_i}(s_0)\big)_{i=0}^{n}$. For $B \subseteq E$, $\mathrm{Reach}^B_F(s_0) = \{F^u(s_0) : u \in B^*\}$.\upshape
\end{thmbox}
\begin{thmbox}\phantomsection\label{lem-argmax}\noindent\textbf{Lemma 1} (Margins decide the argmax).\enspace\itshape Let $O$ be a finite set, $z, \tilde z \in \mathbb R^O$, $y \in O$ and $\tau \in \mathbb R$. If $z_a + 2\tau < z_y$ for every $a \in O \setminus \{y\}$ and $|\tilde z_a - z_a| \le \tau$ for every $a \in O$, then every maximiser of $\tilde z$ over $O$ is $y$. In particular, if $m(v) > 0$, then $f(v) = T(v)$ under every tie rule.\upshape
\end{thmbox}
\begin{proofbox}\noindent\textit{Proof.}\enspace Let $s$ maximise $\tilde z$ over $O$ and suppose $s \neq y$. Then $\tilde z_s \le z_s + \tau < z_y - \tau \le \tilde z_y$, which contradicts $\tilde z_y \le \tilde z_s$. For the second statement take, at each output position $k$, $O = \Sigma$, $y = T_k(v)$, $\tilde z = z = z_k(v)$ and $\tau = 0$; the hypothesis is that the $k$-th term in (\hyperref[eq-margin]{2}) is positive.
\end{proofbox}
\begin{thmbox}\phantomsection\label{lem-regen}\noindent\textbf{Lemma 2} (Regeneration).\enspace\itshape Let $F, \delta\colon S \times E \to S$, $s_0 \in S$ and $B \subseteq E$. Then $F^w(s_0) = \delta^w(s_0)$ for every $w \in B^*$ if and only if $F(s, e) = \delta(s, e)$ for every $s \in \mathrm{Reach}^B_\delta(s_0)$ and $e \in B$. Both conditions are equivalent to equality of the traces of $F$ and $\delta$ on every $w \in B^*$.\upshape
\end{thmbox}
\begin{proofbox}\noindent\textit{Proof.}\enspace ($\Leftarrow$) Induction on the length of $w$. For $w = \varepsilon$ both sides are $s_0$. For $w = ue$ with $u \in B^*$ and $e \in B$,
\[
F^{ue}(s_0) = F\big(F^u(s_0), e\big) = F\big(\delta^u(s_0), e\big) = \delta\big(\delta^u(s_0), e\big) = \delta^{ue}(s_0),
\]
by the induction hypothesis and then by the assumption at $s = \delta^u(s_0) \in \mathrm{Reach}^B_\delta(s_0)$.

($\Rightarrow$) Let $s = \delta^u(s_0)$ with $u \in B^*$, and let $e \in B$. Applying the assumption to $u$ and to $ue$,
\[
F(s, e) = F\big(F^u(s_0), e\big) = F^{ue}(s_0) = \delta^{ue}(s_0) = \delta(s, e).
\]
Traces: the $i$-th entry of the trace on $w$ is the fold over the prefix $e_1 \cdots e_i \in B^*$, so equal folds on all words give equal traces; conversely, the last entries of equal traces are the folds over $w$.
\end{proofbox}
\begin{thmbox}\phantomsection\label{lem-hist}\noindent\textbf{Lemma 3} (Steps that read the history).\enspace\itshape Let $G\colon E^* \times S \times E \to S$, and for $w = e_1 \cdots e_n$ let $s'_0 = s_0$ and $s'_i = G(e_1 \cdots e_{i-1}, s'_{i-1}, e_i)$. If $G(u, s, e) = \delta(s, e)$ for all $u \in B^*$, $s \in \mathrm{Reach}^B_\delta(s_0)$ and $e \in B$, then $s'_n = \delta^w(s_0)$ for every $w \in B^*$.\upshape
\end{thmbox}
\begin{proofbox}\noindent\textit{Proof.}\enspace By induction on $i$, $s'_i = \delta^{e_1 \cdots e_i}(s_0)$. This holds for $i = 0$. If it holds for $i - 1$, then $s'_{i-1} \in \mathrm{Reach}^B_\delta(s_0)$, and the assumption with $u = e_1 \cdots e_{i-1}$ gives $s'_i = \delta(s'_{i-1}, e_i) = \delta^{e_1 \cdots e_i}(s_0)$. Lean proves the lemma with any set $Q \ni s_0$ closed under $\delta$ on $B$ in place of $\mathrm{Reach}^B_\delta(s_0)$, which is the smallest such set.
\end{proofbox}
\begin{thmbox}\phantomsection\label{def-decoder}\noindent\textbf{Definition 3} (Causal decoder).\enspace\itshape A causal decoder of depth $L$ maps tokens $x_0 x_1 \cdots$ to states $h^0_p = \mathrm{emb}(x_p, p)$ and $h^{\ell+1}_p = \Lambda_\ell(p, h^\ell)$, where $h^\ell = (h^\ell_0, h^\ell_1, \dots)$ is the sequence of layer-$\ell$ states and $\Lambda_\ell(p, h)$ depends only on $h_0, \dots, h_p$; the logits of a sequence of length $N$ are $\mathrm{head}(h^L_{N-1})$. A cached forward of a block $b$ after a prefix $\Pi$ reads $h^\ell_p$ for $p < |\Pi|$ from a cache computed on $\Pi$ alone and computes the positions $p \ge |\Pi|$ by the same recursion.\upshape
\end{thmbox}
\begin{thmbox}\phantomsection\label{lem-cache}\noindent\textbf{Lemma 4} (Causality and caching).\enspace\itshape (i) $h^\ell_p$ depends only on $x_0, \dots, x_p$. (ii) The cached forward of $b$ after $\Pi$ computes $h^\ell_p(\Pi b)$ at every layer and position, and hence the logits of $\Pi b$. (iii) Decoding a two-token block $b_1 b_2$ one token at a time on the cache of $\Pi$ gives the logits of $\Pi b_1 b_2$.\upshape
\end{thmbox}
\begin{proofbox}\noindent\textit{Proof.}\enspace (i) Induction on $\ell$: $h^0_p$ reads only $x_p$; if $x$ and $x'$ agree up to $p$, the induction hypothesis gives $h^\ell_i(x) = h^\ell_i(x')$ for $i \le p$, so $h^{\ell+1}_p(x) = h^{\ell+1}_p(x')$ by causality of $\Lambda_\ell$. (ii) Induction on $\ell$. For $p < |\Pi|$ the cache holds $h^\ell_p(\Pi)$, which equals $h^\ell_p(\Pi b)$ by (i). For $p \ge |\Pi|$, layer 0 embeds $b_{p - |\Pi|} = (\Pi b)_p$, and layer $\ell + 1$ applies $\Lambda_\ell(p, \cdot)$ to states equal to $h^\ell(\Pi b)$ by the induction hypothesis. (iii) Decoding $b_1$ on the cache of $\Pi$ leaves the cache of $\Pi b_1$ by (ii), and decoding $b_2$ on that cache gives the logits of $\Pi b_1 b_2$, again by (ii).
\end{proofbox}
\subsection*{Theorem 1}
\begin{proofbox}\noindent\textit{Proof of Theorem 1.}\enspace Fix $c_0 \in I$ and apply Lemma \hyperref[lem-regen]{2} with $S = C$, a single event $E = B = \{\ast\}$, $F(c, \ast) = \sigma_f(c)$ and $\delta(c, \ast) = \sigma_T(c)$. The words over $B$ are $\ast^n$, $F^{\ast^n}(c_0) = \sigma_f^n(c_0)$, the trace on $\ast^n$ is the run of length $n$, and $\mathrm{Reach}^B_\delta(c_0) = \{\sigma_T^n(c_0) : n \ge 0\} = \mathrm{Reach}_T(\{c_0\})$ (Definition \hyperref[def-machine]{1}). By the lemma, the model's run from $c_0$ equals the reference run at every length if and only if
\renewcommand{\theHequation}{eq-agree}
\begin{equation}
\sigma_f(c) = \sigma_T(c) \quad \text{for every } c \in \mathrm{Reach}_T(\{c_0\}).
\tag{13}\label{eq-agree}
\end{equation}

\emph{Sufficiency.} Let $f = T$ on $R_T(I)$ and $c \in \mathrm{Reach}_T(\{c_0\})$. If $\mathrm{view}(c) = \bot$, then $\sigma_f(c) = c = \sigma_T(c)$. Otherwise $v = \mathrm{view}(c) \in R_T(I)$ and $\sigma_f(c) = \mathrm{paint}(c, f(v)) = \mathrm{paint}(c, T(v)) = \sigma_T(c)$. So (\hyperref[eq-agree]{13}) holds for every $c_0 \in I$.

\emph{Necessity.} Let $v \in R_T(I)$, so that $v = \mathrm{view}(c)$ for some $c \in \mathrm{Reach}_T(\{c_0\})$ and $c_0 \in I$; then $\mathrm{view}(c) \neq \bot$, because $R_T(I)$ excludes $\bot$. By (\hyperref[eq-agree]{13}), $\mathrm{paint}(c, f(v)) = \mathrm{paint}(c, T(v))$, and injectivity of $\mathrm{paint}(c, \cdot)$ gives $f(v) = T(v)$.

\emph{The certificate.} If $R_T(I) \subseteq V$ and $m(v) > 0$ for all $v \in V$, Lemma \hyperref[lem-argmax]{1} gives $f = T$ on $V \supseteq R_T(I)$, and sufficiency, which did not use injectivity, applies.
\end{proofbox}
\subsection*{Theorem 2}
Let $\Sigma$ be a set of tokens, $\Pi$ a fixed prompt and $\rho\colon \Sigma^* \to \Sigma$ a step map that returns a maximiser of the logits at the last position of a context. On events $e_1 e_2 \cdots$ from a state $s_0$, pinned decoding computes $s_i = \rho(\Pi\, s_{i-1}\, e_i)$; unpinned decoding keeps the whole context, $x_0 = \Pi\, s_0$, $s_i = \rho(x_{i-1}\, e_i)$ and $x_i = x_{i-1}\, e_i\, s_i$. A row $(s, e)$ of a table $\delta$ passes if the logits at $\Pi\, s\, e$ have their strict maximum at $\delta(s, e)$.

\begin{thmbox}\phantomsection\label{def-counter}\noindent\textbf{Definition 4} (The counterexample).\enspace\itshape The tokens are $\Sigma = \{\omega, \mathsf u, \mathsf v, \mathsf a\}$, where $\omega$ is a sink token, and the prompt is $\Pi = \omega\, \mathsf u\, \mathsf a\, \mathsf u\, \mathsf v\, \mathsf a\, \mathsf v$. The table has states $\{\mathsf u, \mathsf v\}$, the event $\mathsf a$, $\delta(\mathsf u, \mathsf a) = \mathsf u$, $\delta(\mathsf v, \mathsf a) = \mathsf v$ and start $\mathsf u$, so its state is $\mathsf u$ after every input. For $R > 0$ and $c \in \mathbb R$, the logits at the last position of a context $x = x_0 x_1 \cdots x_{N-1}$ are
\renewcommand{\theHequation}{eq-counter}
\begin{equation}
\ell(x)_y = R\, [\, y = x_{N-2} \,] + [\, y = \mathsf v \,]\, \eta(x), \qquad \eta(x) = \frac{\sum_{j=0}^{N-1} e^{\kappa_j} \beta_j}{\sum_{j=0}^{N-1} e^{\kappa_j}},
\tag{14}\label{eq-counter}
\end{equation}
where $[P]$ is 1 if $P$ holds and 0 otherwise, with scores $\kappa_0 = c$ and $\kappa_j = 0$ for $j \ge 1$, and values $\beta_j = 4R$ if $x_j = \mathsf a$ and $\beta_j = 0$ otherwise. The first term is a lookup that returns the state token of the last row, which is the correct output for $\delta$; the second is one attention head at the last position, with score $c$ on the sink, which writes its value to the logit of $\mathsf v$. If $x_0 = \omega$ and $k$ of the $n = N - 1$ tokens after it are $\mathsf a$,
\renewcommand{\theHequation}{eq-write}
\begin{equation}
\eta(x) = \frac{4Rk}{e^{c} + n}.
\tag{15}\label{eq-write}
\end{equation}\upshape
\end{thmbox}
\begin{thmbox}\noindent\textbf{Theorem 2} (Necessity of the pinned-sheet policy, precise form).\enspace\itshape Let $\mathcal T$ be a finite set of contexts, $M$ the largest length of a context in $\mathcal T$ ($M = 0$ if $\mathcal T$ is empty), and $e^{c} = 3M + 5$. For every $R > 0$ and every tie rule, the decoder (\hyperref[eq-counter]{14}): (a) passes both rows of $\delta$ with positive margins; (b) at every context in $\mathcal T$ of the form $\Pi\, w\, s\, \mathsf a$ with $s \in \{\mathsf u, \mathsf v\}$, has its strict maximum at $\delta(s, \mathsf a)$; (c) under unpinned decoding of $\mathsf a^{2M+4}$ ends in the state $\mathsf v$, whereas $\delta$ ends in $\mathsf u$; and (d) under pinned decoding equals $\delta$ on every input.\upshape
\end{thmbox}
\begin{proofbox}\noindent\textit{Proof.}\enspace \emph{Two facts.} At a context ending in $\mathsf v\, \mathsf a$, $\ell_{\mathsf v} = R + \eta \ge R > 0$ while every other logit is 0, since $\eta \ge 0$; so $\mathsf v$ is the strict maximiser. At a context ending in $\mathsf u\, \mathsf a$, $\ell_{\mathsf u} = R$, $\ell_{\mathsf v} = \eta$ and $\ell_\omega = \ell_{\mathsf a} = 0$; so $\mathsf u$ is the strict maximiser if $\eta < R$, and $\mathsf v$ is if $\eta > R$.

(a) A row $\Pi\, s\, \mathsf a$ has $n = 8$ tokens after the sink, three of them $\mathsf a$, so $\eta = 12R/(e^{c} + 8)$ by (\hyperref[eq-write]{15}), which is less than $R$ because $e^{c} = 3M + 5 > 4$. Both facts then give the rows.

(b) A test context $x = \Pi\, w\, \mathsf u\, \mathsf a$ of length $N \le M$ has $n = N - 1$ tokens after the sink and at most $N - 1$ of them $\mathsf a$, so
\[
\eta(x) \le \frac{4R(N - 1)}{3M + 4 + N},
\]
and the right-hand side is less than $R$ if and only if $3N < 3M + 8$, which holds because $N \le M$. Test contexts ending in $\mathsf v\, \mathsf a$ are covered by the first fact.

(c) A state $\mathsf v$ is followed only by $\mathsf v$, by the first fact, so if $s_{i-1} = \mathsf u$, then $s_0 = \cdots = s_{i-1} = \mathsf u$ and the context that determines $s_i$ is $\Pi\, \mathsf u\, (\mathsf a\, \mathsf u)^{i-1}\, \mathsf a$, with $2i + 6$ tokens after the sink, $i + 2$ of them $\mathsf a$. By (\hyperref[eq-write]{15}),
\[
\eta = \frac{4R(i + 2)}{e^{c} + 2i + 6} > R \iff 2i > e^{c} - 2,
\]
and $\eta < R$ if $2i < e^{c} - 2$. Induction on $i$ gives $s_i = \mathsf u$ while $2i < e^{c} - 2$. If $M$ is odd, the step $i = (3M + 3)/2$ has $2i = e^{c} - 2$, so $\eta = R$ and $\ell_{\mathsf u} = \ell_{\mathsf v}$; the tie rule may return either state. In every case, $s_i = \mathsf v$ for the first $i$ with $2i > e^{c} - 2$, if not earlier, and $\mathsf v$ persists by the first fact. For $i = 2M + 4$, $2i = 4M + 8 > 3M + 3 = e^{c} - 2$, so $s_{2M+4} = \mathsf v$.

(d) Under pinned decoding every step reads a row $\Pi\, s\, \mathsf a$ with $s \in \{\mathsf u, \mathsf v\}$, which returns $\delta(s, \mathsf a) \in \{\mathsf u, \mathsf v\}$ by (a). Lemma \hyperref[lem-regen]{2} with $S = \{\mathsf u, \mathsf v\}$, $B = \{\mathsf a\}$ and $F(s, \mathsf a) = \rho(\Pi\, s\, \mathsf a)$ gives equality with $\delta$ on every input.
\end{proofbox}
With $e^{c} = 2N + 5$ instead, the same computation gives a decoder that passes both rows and is exact under unpinned decoding on every input of length at most $N$, yet wrong on $\mathsf a^{N+2}$ (\texttt{no\_\allowbreak{}finite\_\allowbreak{}check}); testing every input up to a given length does not suffice either. Conversely, if the context moves every logit by at most $\tau$ from the logit at the pinned row and every certified margin exceeds $2\tau$, unpinned decoding is exact on every input (\texttt{unpinned\_\allowbreak{}exact\_\allowbreak{}of\_\allowbreak{}close}). In the construction, the head's write is such a perturbation, and it grows without bound.

\subsection*{Theorem 3}
\begin{proofbox}\noindent\textit{Proof of Theorem 3.}\enspace The first statement is Lemma \hyperref[lem-argmax]{1}. For the second, fix $c_0 \in I$ and let the execution path's step $t$ be $G_t(c) = \mathrm{paint}(c, \tilde f_t(\mathrm{view}(c)))$ if $\mathrm{view}(c) \neq \bot$ and $G_t(c) = c$ otherwise, where $\tilde f_t(v)$ is the argmax of the execution-path logits $\tilde z^{(t)}_k(v)$ at each position; these may depend on $t$, and through it on the rest of the batch. Let $c \in \mathrm{Reach}_T(\{c_0\})$ with $v = \mathrm{view}(c) \neq \bot$. Then $v \in R_T(I) \subseteq V$, so $\lVert \tilde z^{(t)}_k(v) - z_k(v) \rVert_\infty \le \tau$ and $m(v) > 2\tau$, and Lemma \hyperref[lem-argmax]{1} at each position gives $\tilde f_t(v) = T(v)$, hence $G_t(c) = \sigma_T(c)$; this also holds when $\mathrm{view}(c) = \bot$. Lemma \hyperref[lem-hist]{3} with a single event, so that the history is the step count, gives $G_t \circ \cdots \circ G_1(c_0) = \sigma_T^t(c_0)$ for every $t$.

In Lean, the first statement is \texttt{sel\_\allowbreak{}eq\_\allowbreak{}of\_\allowbreak{}close} and Lemma \hyperref[lem-hist]{3} is \texttt{histRun\_\allowbreak{}eq\_\allowbreak{}of\_\allowbreak{}closed}; their combination is \texttt{Sketchpad.\allowbreak{}sheet\_\allowbreak{}exact\_\allowbreak{}of\_\allowbreak{}close} for the sheet machine, with the margin condition stated per output position, \texttt{regeneration\_\allowbreak{}of\_\allowbreak{}close} for the token form and \texttt{unpinned\_\allowbreak{}exact\_\allowbreak{}of\_\allowbreak{}close} for unpinned decoding. By Lemma \hyperref[lem-cache]{4}, a pinned step computed on a cache of a fixed prefix, or decoded token by token on it, yields exactly the logits of the full prompt (\texttt{cachedForward\_\allowbreak{}eq}, \texttt{policy\_\allowbreak{}tick\_\allowbreak{}eq}), so in exact arithmetic it is an execution path with $\tau = 0$.
\end{proofbox}
\subsection*{Theorem 4}
\begin{thmbox}\phantomsection\label{def-transcript}\noindent\textbf{Definition 5} (Transcript).\enspace\itshape A transcript makes $K$ calls. Call $i$ receives the input $x_i = q_i(r_{<i})$ and returns $r_i = E_i(x_i)$, where $E_i$ is the procedure that executes it and $r_{<i} = (r_1, \dots, r_{i-1})$; the answer $g(r)$ reads only $r = (r_1, \dots, r_K)$. Call $i$ has a table $T_i$, the function its procedure is certified to compute. The intended computation has inputs $q^*_i$, functions $T^*_i$ and combining program $g^*$, with results $r^*_i = T^*_i(q^*_i(r^*_{<i}))$ and answer $g^*(r^*)$.\upshape
\end{thmbox}
\begin{thmbox}\noindent\textbf{Theorem 4} (Composition, precise form).\enspace\itshape (a) If $g(r) \neq g^*(r^*)$, then for some $i \le K$, $E_i(x_i) \neq T_i(x_i)$, or $T_i(q^*_i(r^*_{<i})) \neq T^*_i(q^*_i(r^*_{<i}))$, or $q_i(r^*_{<i}) \neq q^*_i(r^*_{<i})$; or else $g(r^*) \neq g^*(r^*)$. (b) Let every quantity depend on a transcript $\omega$ in a set $\Omega$ with a measure $\mu$. If $E_i(x_i) = T_i(x_i)$ for every call of every transcript, then
\[
\begin{aligned}
\mu\{g(r) \neq g^*(r^*)\} \le {} & \mu\{\exists i\colon T_i(q^*_i(r^*_{<i})) \neq T^*_i(q^*_i(r^*_{<i}))\} \\
& + \mu\{\exists i\colon q_i(r^*_{<i}) \neq q^*_i(r^*_{<i})\} + \mu\{g(r^*) \neq g^*(r^*)\}.
\end{aligned}
\]\upshape
\end{thmbox}
\begin{proofbox}\noindent\textit{Proof.}\enspace (a) Suppose that none of the four events occurs. We show $r_i = r^*_i$ for all $i \le K$ by strong induction on $i$. If $r_j = r^*_j$ for all $j < i$, then $r_{<i} = r^*_{<i}$, and
\[
r_i = E_i(q_i(r_{<i})) = T_i(q_i(r_{<i})) = T_i(q_i(r^*_{<i})) = T_i(q^*_i(r^*_{<i})) = T^*_i(q^*_i(r^*_{<i})) = r^*_i,
\]
using in turn that the procedure, the transcription and the table are correct. Then $g(r) = g(r^*) = g^*(r^*)$, because $g$ reads only the results and the combining program is correct, a contradiction.

(b) Under the hypothesis, (a) places the event $\{g(r) \neq g^*(r^*)\}$ inside the union of the other three, and the bound follows from monotonicity and subadditivity of $\mu$, which hold for arbitrary sets, so no measurability is needed; a probability distribution gives (\hyperref[eq-compose]{5}). For a procedure certified on the views it reaches, the hypothesis $E_i(x_i) = T_i(x_i)$ is Theorem \hyperref[thm-p]{1} applied to its run on $x_i$, whatever the length of $x_i$ and the order of the calls. In Lean, (a) is proved for arbitrary $E_i$, $T_i$ and $T^*_i$, and (b) for calls executed by sheet procedures (\texttt{Sketchpad.\allowbreak{}sheet\_\allowbreak{}error\_\allowbreak{}bound}) and by the token-form procedure (\texttt{error\_\allowbreak{}bound}), with the hypothesis derived from Theorem \hyperref[thm-p]{1} in each form (\texttt{Sketchpad.\allowbreak{}sheet\_\allowbreak{}compose\_\allowbreak{}exact}, \texttt{compose\_\allowbreak{}exact}).
\end{proofbox}
\section{Certificate format}\label{app-cert}
A certificate is a JSON file distributed with the model graph. It specifies the prompt forms under which it was measured, the numerical precision of each graph, the library version, hashes of the glyph set and of every model file, the device and the time, and, for each table, the number of rows, the number passed and every row's margin in a fixed order, to $10^{-3}$ logits. A client accepts a certificate only if every hash matches the loaded files, and recomputes the lowest-margin tenth of each table before use.

\section{Reproducibility}\label{app-repro}
Every number in this paper is listed, with its source record, in \texttt{paper/\allowbreak{}numbers.\allowbreak{}md} of the Sketchpad repository. Training, server certification, the gate identity test and dialogue fine-tuning are scripted for Modal; the device certificate and browser test run in the client; the algebra tests and cost tables run with \texttt{pnpm vitest run src/\allowbreak{}lib/\allowbreak{}sketchpad/\allowbreak{}program}; the comparison is in \texttt{paper/\allowbreak{}baselines/\allowbreak{}}; the Lean proofs build with \texttt{lake build} in \texttt{paper/\allowbreak{}lean/\allowbreak{}}, and \texttt{lake env lean Paper.\allowbreak{}lean}, run in the same folder, prints the axioms of every statement.

\section*{References}
\begingroup\small
\renewcommand{\refname}{}

\endgroup
\end{document}